\documentclass[letterpaper,journal]{IEEEtran}
\usepackage{amsmath,amsfonts,amssymb}
\usepackage{nicefrac}  
\usepackage{fix-cm}

\usepackage{algorithm}
\usepackage{algorithmic}  
\usepackage{xr}
\usepackage{graphicx}
\usepackage[caption=false,font=normalsize,labelfont=sf,textfont=sf]{subfig}
\usepackage{array}
\usepackage{booktabs}       
\usepackage{tabularx}
\usepackage{multirow}
\usepackage{stfloats}       
\usepackage{placeins}       
\usepackage{float}          

\usepackage{textcomp}
\usepackage{verbatim}
\usepackage{microtype}      
\usepackage{cite}
\usepackage{xcolor}
\usepackage[most]{tcolorbox} 

\usepackage{url}
\usepackage{hyperref}

\newcommand{\upred}{\textcolor{green}{$\uparrow$}}
\newcommand{\downgreen}{\textcolor{red}{$\downarrow$}}

\definecolor{rank1green}{RGB}{0, 180, 0}
\definecolor{rank2blue}{RGB}{46, 139, 192}
\definecolor{rank3orange}{RGB}{229, 114, 0}

\newcommand{\first}[1]{\textcolor{rank1green}{\textbf{#1}}}
\newcommand{\second}[1]{\textcolor{rank2blue}{#1}}
\newcommand{\third}[1]{\textcolor{rank3orange}{#1}}

\newtcolorbox{observationbox}[1]{
  colback=gray!5!white, 
  colframe=black,       
  fonttitle=\bfseries,
  arc=5pt,              
  outer arc=5pt,
  boxrule=1pt,          
  title=#1,             
  coltitle=black,       
  attach title to upper, 
  after title={\hspace{0.5em}}, 
  left=5pt, right=5pt, top=5pt, bottom=5pt 
}

\begin{document}

\title{EEG-FM-Audit: A Systematic Evaluation and Analysis Pipeline for EEG Foundation Models}


\author{Xianheng~Wang, Yige~Yang, and~Damien~Coyle,~\IEEEmembership{Senior~Member,~IEEE}%
\thanks{Xianheng Wang and Damien Coyle are with the Bath Institute for the Augmented Human, University of Bath, Bath BA2 7AY, U.K. (e-mail: wangxianh940621@gmail.com; dhc30@bath.ac.uk).}%
\thanks{Yige Yang is with the Department of Computer Science, University of Bath, Bath BA2 7AY, U.K. (e-mail: yy2541@bath.ac.uk).}}





\maketitle

\begin{abstract}
EEG foundation models (EEG-FMs) show potential for learning generalized representations from heterogeneous EEG signals. However, existing studies exhibit three critical limitations: opaque model tuning, unverified contributions of complex learning paradigms to EEG-FMs, and a limited understanding of the neurophysiological features that FMs rely on to make decisions. To address these challenges, we propose \textbf{EEG-FM-Audit}, a systematic evaluation and diagnostic pipeline comprising three modules: (1) a fairness-calibrated benchmarking protocol ensuring fair comparisons through model-tailored hyperparameter optimization for EEG-FMs and supervised baselines; (2) three paradigm-level ablation
workflows isolating contributions of heterogeneous data
scale, self-supervised pretraining, and pretrained linguistic priors; and (3) a neurophysiological probing (NPP) framework exploring whether FMs leverage valid EEG features. Extensive experiments across four leading EEG-FMs and five supervised models on four datasets reveal that well-tuned supervised baselines can match or outperform tuned FMs in some cases, despite requiring significantly fewer parameters. Furthermore, ablation analysis indicates that FM pre-training is indispensable for data-scarce downstream tasks, whereas the contribution of pre-trained linguistic components depends heavily on model architecture. Finally, NPP analysis shows that FMs generally extract the established temporal, spatial, and spectral characteristics of the evaluated datasets, establishing a systematic diagnostic benchmark for assessing EEG-FMs.
\end{abstract}


\section{Introduction}
\IEEEPARstart{A}{s} a technique for measuring electrical activity in the brain, electroencephalography (EEG) has been widely used in research and in commercial and clinical applications. Accurate classification of EEG signals is crucial because it enables effective control in brain–computer interfaces (BCIs)~\cite{blankertz2001classifying} and facilitates a deeper understanding of brain function. In recent years, deep learning (DL)-based methods have dominated the field of EEG classification, especially for supervised models~\cite{Lawhern2018EEGNet, Li2021TSSEFFNet, wang2024indepth}. While supervised methods have demonstrated competitive performance, accurate EEG classification remains challenging due to the inherent non-stationarity and low signal-to-noise ratio of the data.  

To enhance generalization and robustness, recent studies~\cite{cui2024neurogpt,jiang2024large,wang2024eegpt,jiang2025neurolm} have pivoted toward large-scale EEG Foundation Models (FMs), which leverage massive and heterogeneous datasets to capture generalized neural representations. The reported results~\cite{wang2024eegpt, lee2025are} demonstrate their strong potential for decoding EEG signals. However, the lack of systematic evaluation and analysis hinders a deeper understanding of EEG-FMs. Consequently, several recent studies~\cite{jiang2025neurolm, xiong2026eegfmbench, kontras2026neuroatlas} have turned their attention to comprehensively evaluating EEG-FM performance and analyzing their architectural designs. Although these studies have gained useful insights into EEG-FMs, they generally exhibit three limitations. The first limitation is the opaque tuning methods for both baselines and EEG-FMs. Existing studies~\cite{jiang2025neurolm, xiong2026eegfmbench, kontras2026neuroatlas} generally omit systematic hyperparameter tuning or lack transparent optimization protocols for benchmarking. This can lead to inaccurate and unfair evaluations, as suboptimally tuned models fail to reflect their true capabilities, particularly for trained-from-scratch supervised models (here referred to as 'supervised baselines'), which are typically far more sensitive to hyperparameter selection~\cite{bouthillier2021accounting}. The second limitation is a narrow focus on model layers rather than learning paradigms. Existing studies~\cite{lee2025are, liu2026eeg, xiong2026eegfmbench} often lack analysis of EEG-FM learning paradigms, such as the effects of large-scale pretraining, the entire pretraining step, and so on. An in-depth understanding of paradigm-level contributions is essential for advancing EEG-FM design. The third limitation is insufficient insight into FMs’ reliance on valid neurophysiological features. Specifically, it remains unclear whether EEG-FMs genuinely capture and utilise meaningful temporal, spatial, and spectral features of EEG signals in their decision-making.

\begin{figure*}[ht]
\centering
\includegraphics[height=9cm ,width=43cm,angle=0,scale=0.4]{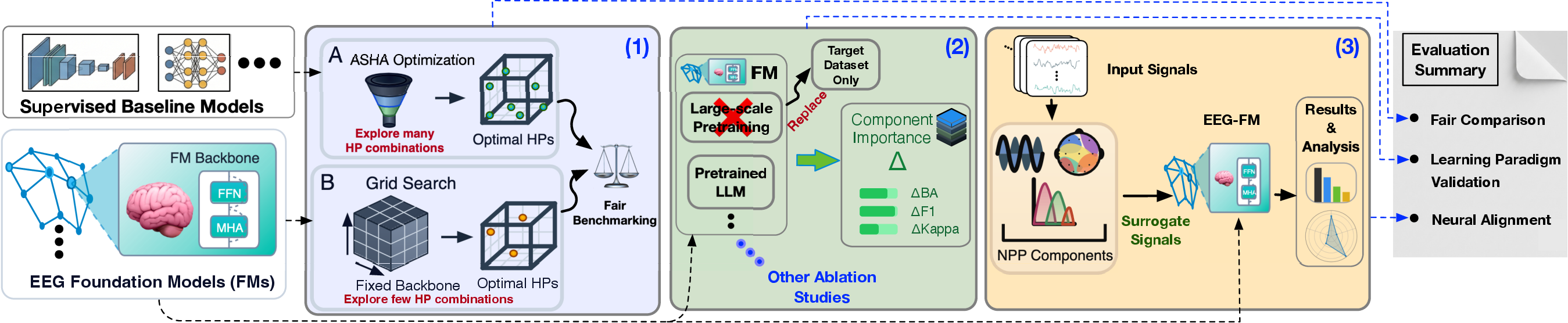}
\caption{Overview of the EEG-FM-Audit framework. A three-stage systematic pipeline for evaluating and analyzing EEG foundation models (FMs): (1) A fairness-calibrated benchmarking protocol establishes fair comparisons between supervised baselines and EEG-FMs. (2) Paradigm-level ablation studies deconstruct core components (e.g., large-scale pretraining and pretrained language backbone) to perform learning paradigm validation; and (3) NPP utilizes surrogate signals to quantify neural alignment between model decision-making and established neuroscientific findings across temporal, spatial, and spectral dimensions. The pipeline produces a comprehensive evaluation summary assessing the EEG-FMs' actual performance gains, the isolated contributions of key EEG-FM core components, and their reliance on biologically meaningful neurophysiological features.}
\label{fig_EEGFMAudit}
\end{figure*}

To address these limitations, we design a comprehensive evaluation and analysis pipeline, namely EEG-FM-Audit (see Fig.~\ref{fig_EEGFMAudit}). First, our fairness-calibrated benchmarking protocol tunes supervised baselines and EEG-FMs using model-tailored strategies to address the existing opaque tuning issue. For lightweight supervised baselines, we adopt a parallel two-stage Asynchronous Successive Halving Algorithm (ASHA)~\cite{li2020system} optimization to efficiently explore diverse architectural and training hyperparameters. Conversely, because running massive concurrent trials is computationally prohibitive for large-scale EEG-FMs and their fixed pretrained architectures narrow the search space, we employ a targeted grid search. Second, we conduct three paradigm-level ablations to examine large-scale pretraining data, the pretraining stage itself, and linguistic backbones. Because these paradigms distinguish EEG-FMs from supervised baselines, isolating their effects directly verifies their actual performance benefits.  Third, we propose an NPP framework to verify whether EEG-FMs rely on valid neurophysiological features across temporal, spatial, and spectral dimensions by using surrogate signals.

To demonstrate the utility of EEG-FM-Audit and provide valuable insights in the field, we utilize it to evaluate four advanced EEG-FMs and five representative supervised EEG baseline models across four popular datasets with different data scales and tasks. In summary, our main contributions are as follows:
\begin{itemize}
    \item We propose and adopt a fairness-calibrated benchmarking protocol using suitable methods to transparently tune supervised baselines and EEG-FMs, establishing fair benchmarking.
    \item We design three paradigm-level ablation studies to explore the contribution of large-scale pretraining, the entire pretraining paradigm, and pretrained linguistic backbone in EEG-FMs.
    \item We introduce NPP as a neuroscientific sanity check for probing EEG-FM signatures. To our knowledge, this is the first systematic analysis of EEG-FMs across temporal, spatial, and spectral dimensions.
    \item We conduct extensive experiments across recent EEG-FMs, supervised baselines, and multiple EEG datasets, and provide a systematic and reproducible pipeline for fair and comprehensive EEG-FM evaluation, with code publicly available at \url{https://github.com/Henrywang621/EEG-FM-Audit}.
\end{itemize}

\section{Related Work}
\label{gen_inst}

\subsection{Evolution of EEG Decoding Models} 

DL-based methods have been dominant for EEG decoding over the past decade. Earlier DL-based works~\cite{schirrmeister2017deep, Lawhern2018EEGNet, Jiang2024CSPNet} often emulate the computational steps of a classic non-DL method, i.e., Filter-Bank Common Spatial Pattern (FBCSP)~\cite{ang2008filter}, when designing their methods. A representative method is EEGNet~\cite{Lawhern2018EEGNet}, which employs depthwise and separable convolutions to extract compact spatiotemporal representations across diverse BCI paradigms. To further improve decoding performance, the subsequent works~\cite{Li2021TSSEFFNet, Zhao2025MSCFormer,Zhao2024CTNet} introduce the attention-mechanism~\cite{wang2024indepth} into their architectures, as the attention-mechanism can enable models to focus on task-relevant features. For example, Li et al.~\cite{Li2021TSSEFFNet} design an attention-based feature fusion block to selectively integrate the extracted spatiotemporal and spectral features. Their experimental results demonstrate that the proposed model outperforms representative earlier deep learning (DL) models~\cite{schirrmeister2017deep, Lawhern2018EEGNet} on two public Motor Imagery EEG datasets. Zhao et al.~\cite{Zhao2024CTNet} first adopt a convolutional module to extract spatiotemporal features. Then, they develop a Transformer-based encoder that uses multi-head attention to capture global dependencies among the extracted features. 

Inspired by the unprecedented success of Large Language Models (LLMs)~\cite{vaswani2017attention, brown2020language}, recent studies present EEG foundation models (EEG-FMs)~\cite{jiang2024large, cui2024neurogpt, jiang2025neurolm}. Unlike traditional supervised models confined to single-dataset training and evaluation, EEG-FMs leverage flexible architectures to pretrain on heterogeneous data across multiple datasets with varying channel configurations and recording lengths. This enables EEG-FMs to learn universal neural representations that can be effectively transferred to a wide range of downstream tasks via fine-tuning. Currently, most EEG-FMs, such as~\cite{cui2024neurogpt, wang2024eegpt, jiang2024large, yang2026are}, adopt a direct classification paradigm that maps extracted features to class probabilities. For example, Cui et al.~\cite{cui2024neurogpt} propose a pioneering EEG-FM called NeuroGPT, which is first pre-trained on
large-scale EEG data using causal self-supervision to reconstruct masked tokens, and then the pre-trained backbone is adapted
for downstream BCI tasks by appending a linear classification head. Wang et al.~\cite{wang2024eegpt} design EEGPT, which adopts dual self-supervised learning to jointly optimize reconstruction and alignment objectives during pretraining and subsequently applies linear probing for downstream tasks. Beyond direct classification, an alternative approach is the classification-via-generation paradigm, which formulates prediction as conditional token generation (e.g., outputting ``Yes/No'' for binary classification). A representative model adopting this strategy is NeuroLM~\cite{jiang2025neurolm}, which integrates a pre-trained LLM with a text-aligned neural tokenizer to bridge brain signals and natural language, leveraging rich linguistic knowledge to enhance the semantic representation of EEG signals. Although existing EEG-FMs consistently report superior performance over supervised baselines, the lack of transparent and rigorous hyperparameter optimization for these baselines undermines the validity of their conclusions. To resolve this ambiguity, we propose EEG-FM-Audit to systematically re-evaluate EEG-FMs and uncover their genuine performance gains through rigorous, fair comparisons.

\subsection{Systematic Gaps in Existing EEG Foundation Model Literature}

The current literature on EEG-FMs can be broadly bifurcated into qualitative surveys~\cite{yao2025foundation, li2025foundation} and empirical benchmarks~\cite{lee2025are, liu2026eeg, xiong2026eegfmbench,kuruppu2025eeg,kontras2026neuroatlas}. The former focuses on establishing conceptual taxonomies and summarizing high-level trends based on original author papers. For example, Li et al.~\cite{li2025foundation} classify and summarize EEG-FMs according to their output modalities and discuss open issues and future research directions. Although these provide comprehensive overviews, they often lack independent experimental verification and analysis. 

Unlike qualitative works, empirical benchmarks involve the active implementation and systematic evaluation of models to provide a data-driven evaluation of their performance, fine-tuning dynamics, generalization capabilities, and other aspects of model behavior. For example, Lee et al.~\cite{lee2025are} evaluate and analyze two advanced EEG-FMs using layer-wise ablation and Low-Rank Adaptation. It is the first work to comprehensively assess fine-tuned performance of the EEG-FMs on multiple downstream tasks. Kuruppu et al.~\cite{kuruppu2025eeg} analyze several EEG-FMs released up to 2024 along three axes: input data, modeling, and evaluation. The authors also thoroughly discuss existing research gaps and provide potential directions for future research. A very recent study~\cite{kontras2026neuroatlas} conducts a systematic evaluation across various clinical and BCI tasks. As the largest current EEG benchmark (NeuroAtlas), it evaluates not only EEG-FMs but also general time-series FMs as baselines and highlights that NeuroAtlas exposes gaps in EEG-FMs and provides the datasets and metrics for the next generation of unified EEG-FMs.

Although existing empirical benchmarks provide broad performance comparisons for EEG-FMs, they exhibit three critical limitations: (1) opaque or omitted hyperparameter optimization; (2) unverified contributions of core EEG-FM learning paradigms to downstream performance; and (3) insufficient verification of whether EEG-FMs rely on valid neurophysiological features. In this paper, we address these gaps by introducing EEG-FM-Audit, a unified evaluation and analysis pipeline consisting of a fairness-calibrated benchmarking protocol, three paradigm-level ablations, and an NPP framework.


\section{EEG-FM-Audit: A Multi-Level Evaluation Pipeline}
\label{EEG_Audit}

In this section, we first introduce our fairness-calibrated benchmarking protocol in Section~\ref{susection_FCBP}, which applies tailored hyperparameter optimization strategies to both supervised baselines and EEG-FMs. Next, Section~\ref{Sec3.2_AB} presents three ablation studies designed to evaluate the effectiveness of key learning paradigms in EEG-FMs. Finally, Section~\ref{Sec3.3NPP} details the workflow of our proposed neurophysiological probing (NPP) framework.

\begin{figure}[ht]
\centering
\includegraphics[height=22cm ,width=17 cm,angle=0,scale=0.4]{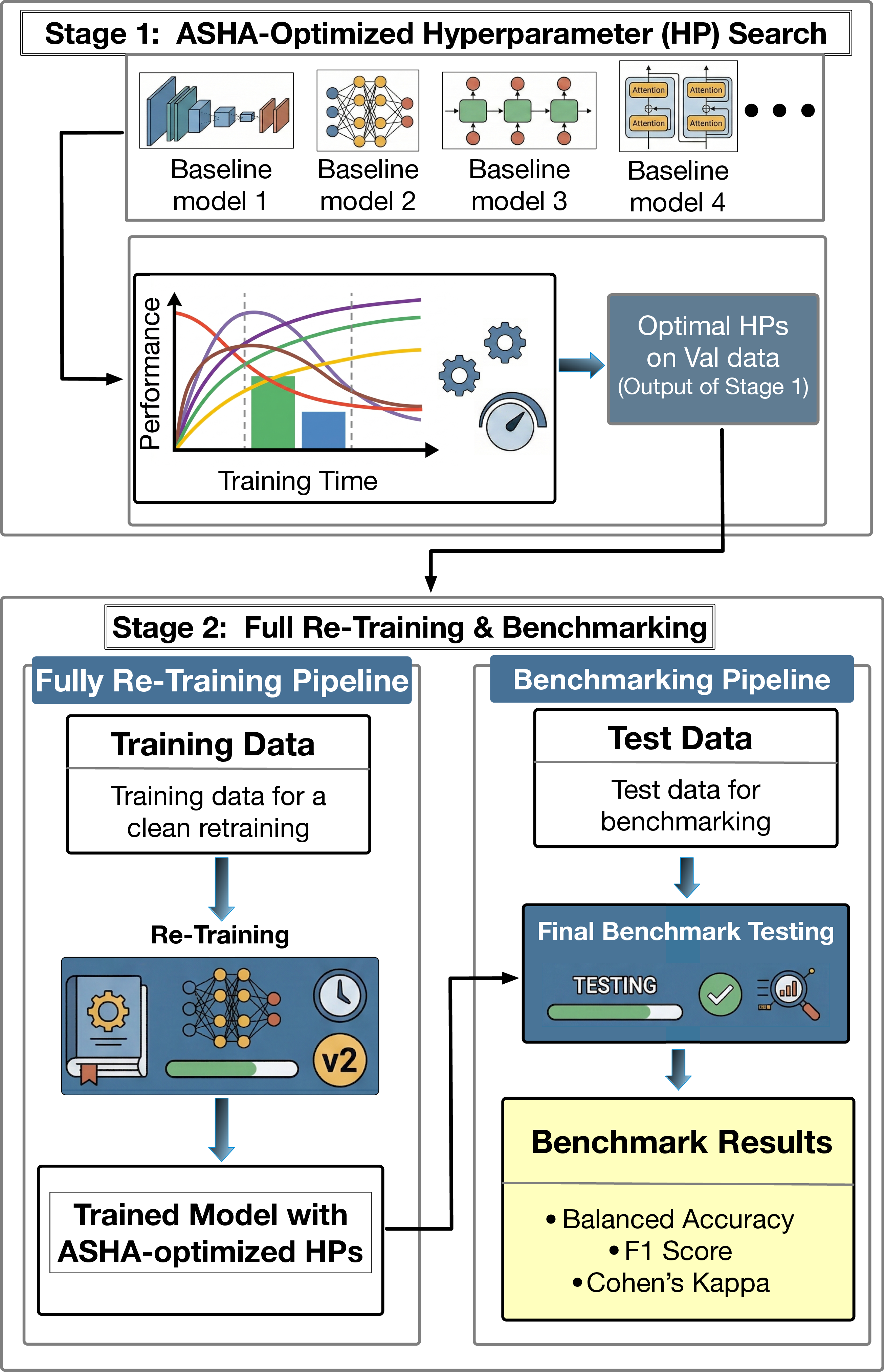}
\caption{Overview of the two-stage ASHA-driven baseline optimization. Stage 1 (HP Search): ASHA explores the hyperparameter space on the training and validation splits to identify the optimal configuration ($\Theta^*$) for each baseline architecture. Stage 2 (Full Re-Training \& Benchmarking): Each baseline is retrained from scratch using $\Theta^*$ and evaluated on the held-out test set across Balanced Accuracy, F1 Score, and Cohen's Kappa.}
\label{fig_ashaproto}
\end{figure}

\begin{algorithm}[!t]
\caption{ASHA-driven baseline benchmarking protocol}
\label{alg:generic_benchmarking}
\begin{algorithmic}[1]
\REQUIRE Model $\mathcal{M}$, Search Space $\Omega$, Data $\mathcal{D}$, Strategy $\Gamma$
\ENSURE Performance Metric $\Phi = \mu \pm \sigma$

\FORALL{iteration $k \in \{1, \dots, K\}$}
    \STATE $\{\mathcal{D}_{tr}, \mathcal{D}_{val}, \mathcal{D}_{ts}\} \gets \text{GetSplit}(\mathcal{D}, \Gamma, k)$ \COMMENT{Static split or $k$-th fold}
    \STATE $\mathcal{M}_{init} \gets \text{InitializeWeights}(\mathcal{M}, \text{seed } k)$ \COMMENT{Stochastic initialization}
    
    \STATE \COMMENT{\textbf{Stage 1: ASHA-Optimized HPs Search}}
    \STATE $\Theta^*_k \gets \text{ASHA-Tune}(\mathcal{M}_{init}, \Omega, \mathcal{D}_{tr}, \mathcal{D}_{val})$ \COMMENT{Optimal HPs search on validation set}
    
    \STATE \COMMENT{\textbf{Stage 2: Full Re-Training \& Benchmarking}}
    \STATE $\mathcal{M}_{ASHA} \gets \text{Retrain}(\mathcal{M}_{init}, \Theta^*_k, \mathcal{D}_{tr} \cup \mathcal{D}_{val})$ \COMMENT{Retrain with the optimal HPs $\Theta^*_k$}
    \STATE $\phi_k \gets \text{Assess}(\mathcal{M}_{ASHA}, \mathcal{D}_{ts})$ \COMMENT{Evaluate on unseen test set}
\ENDFOR

\RETURN $\Phi \gets \text{mean}(\{\phi_k\}) \pm \text{std}(\{\phi_k\})$ \COMMENT{Aggregated results}
\end{algorithmic}
\end{algorithm}

\subsection{Fairness-Calibrated Benchmarking Protocol}
\label{susection_FCBP}

A common flaw in existing EEG-FM studies is the opaque or omitted hyperparameter optimization (HPO) of supervised baselines, which often leads to skewed comparisons. To establish a rigorous evaluation, we propose a fairness-calibrated benchmarking protocol that applies paradigm-specific optimization strategies to properly tune both supervised baselines and EEG-FMs.

Supervised models trained from scratch usually exhibit substantial sensitivity to hyperparameter configurations (see Section~\ref{FMBenchmarking}), with performance varying considerably across architectural and optimization choices (e.g., kernel sizes, hidden dimensions, learning rates, and batch sizes), making careful HPO essential for fair benchmarking. To efficiently explore a wide range of hyperparameter combinations, we employ a two-stage, ASHA-driven baseline optimization protocol (see Fig.~\ref{fig_ashaproto}). As described in Algorithm~\ref{alg:generic_benchmarking}, given a baseline model $\mathcal{M}$, hyperparameter search space $\Omega$, dataset $\mathcal{D}$, and partitioning strategy $\Gamma$ across $K$ runs (e.g., $K=5$ random seeds for fixed holdout or $K=9$ folds for subject-wise cross-validation), each iteration $k \in \{1, \dots, K\}$ splits $\mathcal{D}$ into $\mathcal{D}_{tr}$, $\mathcal{D}_{val}$, and $\mathcal{D}_{ts}$. Stage 1 identifies the optimal configuration $\Theta^*_k \in \Omega$ on $\mathcal{D}_{val}$ via ASHA, which dynamically terminates unpromising trials via early stopping. Stage 2 then retrains $\mathcal{M}$ from scratch using $\Theta^*_k$ and evaluates its performance on the held-out test set $\mathcal{D}_{ts}$ to obtain the iteration score $\phi_k$, reporting final performance across all $K$ runs as $\Phi = \mu \pm \sigma$.

Unlike supervised models, pretrained EEG-FMs typically employ fixed architectural backbones, resulting in a relatively constrained hyperparameter search space. Moreover, EEG-FMs are often substantially larger than supervised models, making large-scale parallel hyperparameter optimization computationally prohibitive. Consequently, our two-stage ASHA-driven optimization strategy, which is designed to efficiently explore large search spaces through massive parallelization, is less suitable for EEG-FMs. Instead, we employ a compact grid search to evaluate the limited set of candidate hyperparameter configurations for EEG-FMs.

\begin{figure*}[t]
\centering
\includegraphics[height=11cm ,width=34cm,angle=0,scale=0.4]{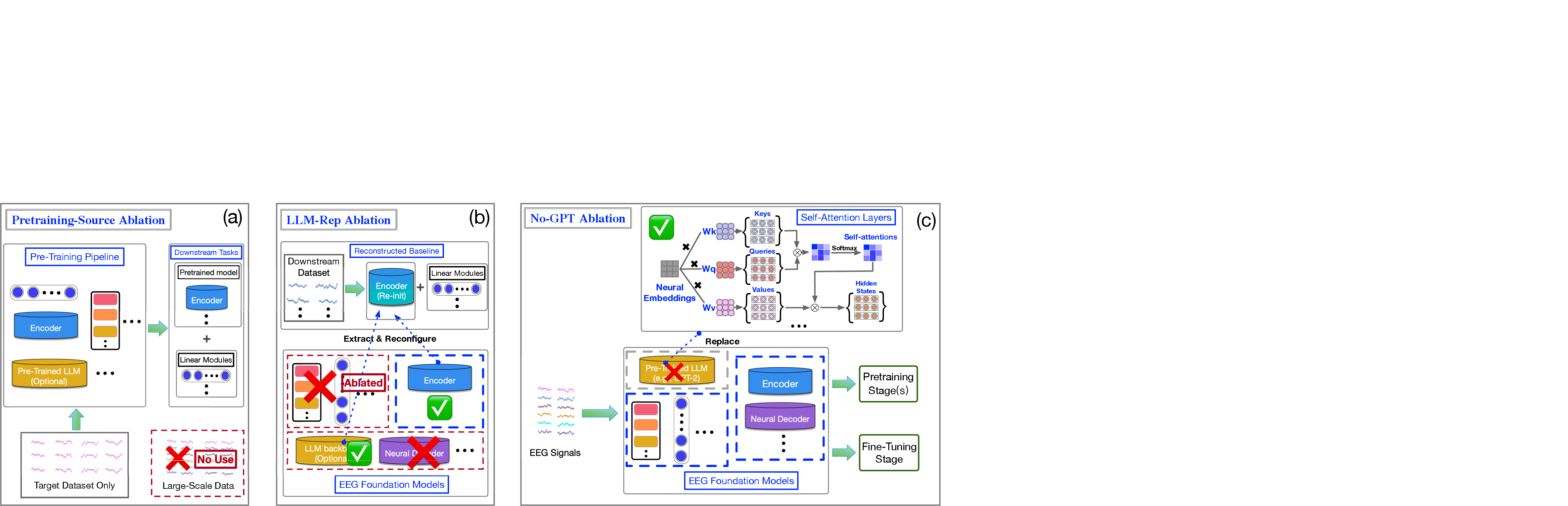}
\caption{Overview of three paradigm-level ablation studies. A systematic deconstruction of foundation model components to isolate their individual contributions: (a) Pretraining-Source Ablation evaluates the necessity of large-scale pretraining versus domain-specific pretraining. (b) LLM-Rep Ablation assesses the impact of the whole pretraining stage by retaining only the foundation model's core transformer backbone and attaching a standard classification head to construct a supervised comparison model. (c) No-GPT Ablation isolates the effect of pretrained linguistic priors by replacing the pretrained LLM with three randomly initialized self-attention layers.}
\label{fig_3AS}
\end{figure*}

\subsection{Paradigm-Level Ablation: Revisiting the Benefits of EEG-FM Learning Paradigms} 
\label{Sec3.2_AB}

While EEG-FMs leverage sophisticated paradigms, such as self-supervised pretraining~\cite{weinzaepfel2022croco} on heterogeneous datasets and the integration of pretrained language model backbones (e.g., GPT-2~\cite{radford2019language}), the individual necessity of these components remains an open question. It is unclear whether the increased complexity introduced by the LLM-inspired paradigms~\cite{minaee2024llm} leads to tangible performance gains. To investigate this question, we design three paradigm-level ablation studies (see Fig.~\ref{fig_3AS} and Table~\ref{tab:paradigm_ablations}) to disentangle the contributions of key learning paradigms within EEG-FMs.

First, pretraining-source ablation (see Fig.~\ref{fig_3AS} (a)) evaluates the effect of large-scale heterogeneous pretraining by comparing the full FM against a baseline pretrained solely on the target dataset (i.e., the downstream training data). Second, LLM-Rep Ablation (see Fig.~\ref{fig_3AS} (b)) isolates the effect of the whole self-supervised paradigm. Specifically, we reconstruct the FM into a traditional supervised framework to determine if performance gains stem from the pretraining stage. Finally, No-GPT Ablation (see Fig.~\ref{fig_3AS} (c)) replaces the pretrained LLM backbone  (e.g., GPT-2~\cite{radford2019language}) with three randomly initialized self-attention layers to explore whether linguistic priors from a pretrained LLM actually contribute to neural dynamics decoding. These studies allow us to evaluate whether the increased complexity of the FM framework leads to tangible performance gains over traditional DL paradigms. 

\begin{table}[!t]
\centering
\caption{Summary of Paradigm-Level Ablation Studies. All resulting models $\mathcal{M}_{final}$ are evaluated on the held-out test set.}
\label{tab:paradigm_ablations}
\renewcommand{\arraystretch}{1.2}
\setlength{\tabcolsep}{3.5pt}
\footnotesize
\begin{tabularx}{\columnwidth}{@{} l >{\raggedright\arraybackslash}X @{}}
\toprule
\textbf{Stage} & \textbf{Formulation \& Description} \\ \midrule

\multicolumn{2}{@{}l@{}}{\textbf{Reference: Default EEG-FM Pipeline}} \\
\hspace{0.3em}Pretraining & $\Theta^*_{multi} {\leftarrow} \mathrm{Pretrain}(\mathcal{M}_{FM}(\Theta_{init}),\allowbreak \mathcal{D}_{multi},\allowbreak \mathcal{L}_{SSL})$ on multi-dataset corpus $\mathcal{D}_{multi}$ with EEG-FM architecture $\mathcal{M}_{FM}$ and self-supervised learning (SSL) loss $\mathcal{L}_{SSL}$. \\
\hspace{0.3em}Downstream & $\mathcal{M}_{final} {\leftarrow} \mathrm{Finetune}(f_{out} {\circ} \mathcal{M}_{FM}(\Theta^*_{multi}),\allowbreak \mathcal{D}^{tr}_{target},\allowbreak \mathcal{L}_{task})$ with readout head $f_{out}$, target training data $\mathcal{D}^{tr}_{target}$, and downstream task loss $\mathcal{L}_{task}$. \\ \midrule

\multicolumn{2}{@{}l@{}}{\textbf{1. Pretraining-Source Ablation}} \\
\hspace{0.3em}Architecture & Retain unmodified $\mathcal{M}_{FM}$ with random weights $\Theta_{init} {\sim} \mathcal{N}(0, \sigma^2)$. \\
\hspace{0.3em}Phase 1 & $\Theta^*_{target} {\leftarrow} \mathrm{Pretrain}(\mathcal{M}_{FM}(\Theta_{init}),\allowbreak \mathcal{D}^{tr}_{target},\allowbreak \mathcal{L}_{SSL})$ restricted strictly to target training data $\mathcal{D}^{tr}_{target}$. \\
\hspace{0.3em}Phase 2 & Attach standard classifier $f_{cls}$ with supervised loss $\mathcal{L}_{sup}$:\newline
$\mathcal{M}_{final} {\leftarrow} \mathrm{Finetune}(f_{cls} {\circ} \mathcal{M}_{FM}(\Theta^*_{target}),\allowbreak \mathcal{D}^{tr}_{target},\allowbreak \mathcal{L}_{sup})$. \\ \midrule

\multicolumn{2}{@{}l@{}}{\textbf{2. LLM-Rep Ablation}} \\
\hspace{0.3em}Architecture & Extract core encoder $\mathcal{M}_{enc}$ from $\mathcal{M}_{FM}$; initialize $\Theta_{init} {\sim} \mathcal{N}(0, \sigma^2)$. \\
\hspace{0.3em}Phase 1 & Omitted completely (no $\mathcal{L}_{SSL}$ optimization or pretraining corpus). \\
\hspace{0.3em}Phase 2 & Attach $f_{cls}$ directly to $\mathcal{M}_{enc}$ and train from scratch:\newline
$\mathcal{M}_{final} {\leftarrow} \mathrm{TrainFromScratch}(f_{cls} {\circ} \mathcal{M}_{enc}(\Theta_{init}),\allowbreak \mathcal{D}^{tr}_{target},\allowbreak \mathcal{L}_{sup})$. \\ \midrule

\multicolumn{2}{@{}l@{}}{\textbf{3. No-GPT Ablation}} \\
\hspace{0.3em}Substitution & Express $\mathcal{M}_{FM} {=} h {\circ} \mathcal{M}_{LLM} {\circ} g$ ($g, h$: preceding/succeeding sub-networks); replace pretrained LLM module $\mathcal{M}_{LLM}$ with random self-attention $\mathcal{M}_{target}$ to form $\mathcal{M}_{sub} {\leftarrow} h {\circ} \mathcal{M}_{target} {\circ} g$. \\
\hspace{0.3em}Phase 1 & $\Theta^*_{sub} {\leftarrow} \mathrm{Pretrain}(\mathcal{M}_{sub},\allowbreak \mathcal{D}_{multi},\allowbreak \mathcal{L}_{SSL})$ on corpus $\mathcal{D}_{multi}$. \\
\hspace{0.3em}Phase 2 & Attach readout module $f_{out}$:\newline
$\mathcal{M}_{final} {\leftarrow} \mathrm{Finetune}(f_{out} {\circ} \mathcal{M}_{sub}(\Theta^*_{sub}),\allowbreak \mathcal{D}^{tr}_{target},\allowbreak \mathcal{L}_{task})$. \\ \bottomrule
\end{tabularx}
\end{table}

\subsection{Neurophysiological Probing: Decoding the Foundation Model Logic}

\label{Sec3.3NPP}

\begin{figure*}[t]
\centering
\includegraphics[height=17cm ,width=34cm,angle=0,scale=0.4]{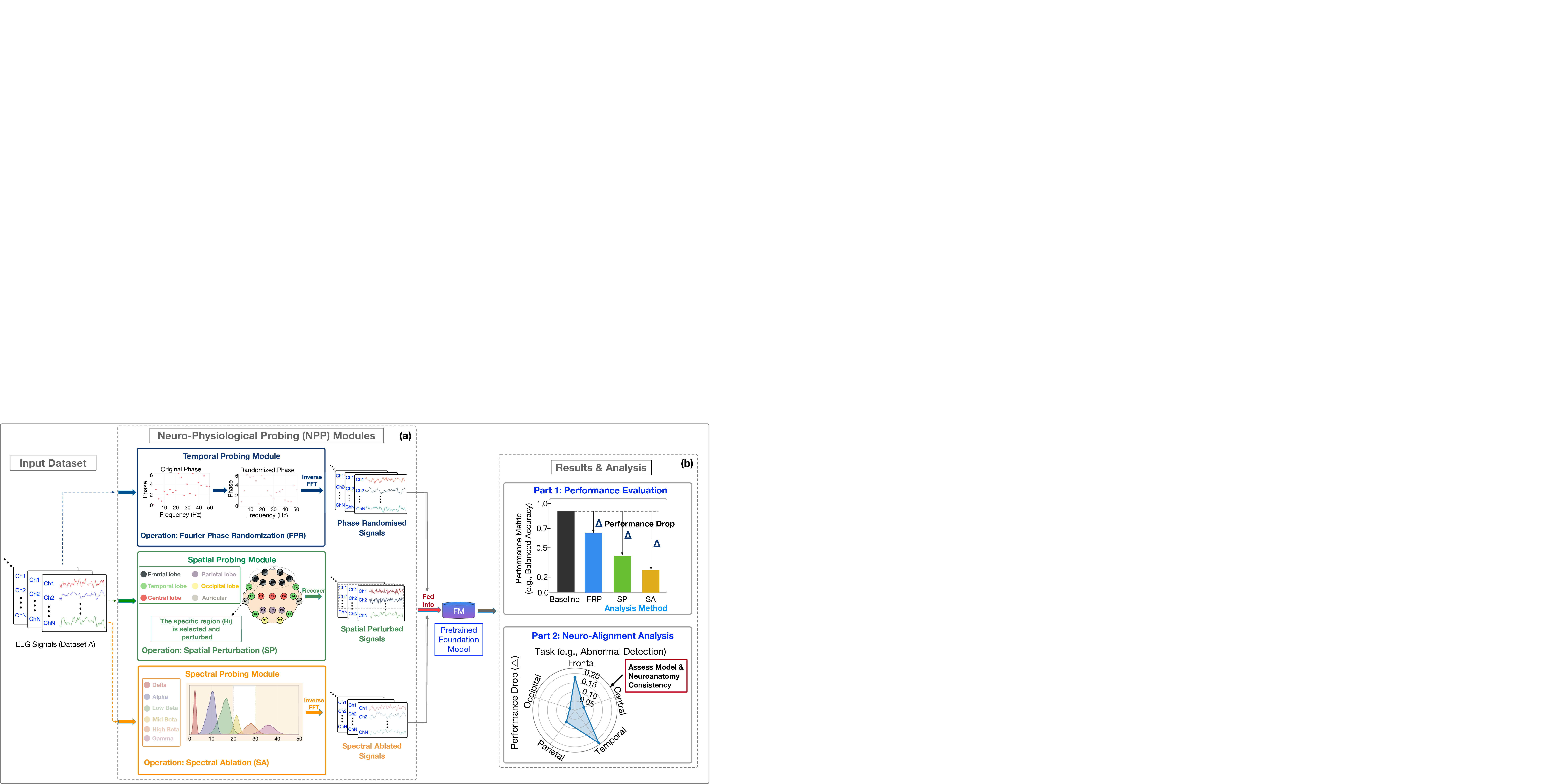}
\caption{Overview of the neurophysiological probing framework. (a) The framework transforms EEG input signals using three modules: Temporal Probing via phase randomization, Spatial Probing via region of interest (ROI) specific perturbation, and Spectral Probing via band-wise ablation. These modules generate reconstructed signals to evaluate model dependencies. (b) Resulting performance degradation is measured to quantify model reliance on specific neurophysiological features (Part 1) and to verify neuroanatomical consistency through alignment analysis (Part 2).}
\label{fig_NPP}
\end{figure*}

\begin{table}[!t]
\centering
\caption{Summary of Neurophysiological Probing (NPP) Modules}
\label{tab:npp_modules}
\renewcommand{\arraystretch}{1.2}
\setlength{\tabcolsep}{3.5pt}
\footnotesize
\begin{tabularx}{\columnwidth}{@{} l >{\raggedright\arraybackslash}X @{}}
\toprule
\textbf{Step} & \textbf{Mathematical Specification ($\mathbf{X}, \hat{\mathbf{X}} \in \mathbb{R}^{C \times T}$)} \\ \midrule

\multicolumn{2}{@{}l@{}}{\textbf{1. Temporal Probing: Fourier Phase Randomization (FPR)}} \\
\hspace{0.3em}Spectrum & Compute demeaned spectrum $\mathbf{Y} {\leftarrow} \mathrm{rFFT}(\mathbf{X} - \bar{\mathbf{X}}) = |\mathbf{Y}|e^{j\phi} \in \mathbb{C}^{C \times F}$ across $F = \lfloor T/2 \rfloor + 1$ bins. \\
\hspace{0.3em}Surrogate & Sample shared phase $\theta {\sim} \mathcal{U}(0, 2\pi)^{1 \times F}$ ($\theta_0 {=} 0$) across all $C$ channels:\newline
$\hat{\mathbf{X}} {\leftarrow} \mathrm{irFFT}(|\mathbf{Y}| {\odot} e^{j\theta},\allowbreak n{=}T) + \bar{\mathbf{X}}$. \\
\hspace{0.3em}Property & Destroys phase-locked temporal dynamics; preserves each channel's power spectrum $|\mathbf{Y}|$ and variance. \\ \midrule

\multicolumn{2}{@{}l@{}}{\textbf{2. Spatial Probing: ROI-Specific Perturbation (SP)}} \\
\hspace{0.3em}Setup & Define binary region mask $\mathbf{M}_r \in \{0,1\}^{C \times 1}$, signal scale $\sigma_{\mathbf{X}} {\leftarrow} \mathrm{std}(\mathbf{X})$, and noise $\mathbf{Z} {\sim} \mathcal{N}(0,1)^{C \times T}$. \\
\hspace{0.3em}Surrogate & Inject scale-aware noise with intensity $\lambda$ into target ROI:\newline
$\hat{\mathbf{X}} {\leftarrow} \mathbf{X} + \lambda \cdot \sigma_{\mathbf{X}} \cdot (\mathbf{M}_r {\odot} \mathbf{Z})$. \\
\hspace{0.3em}Property & Perturbs localized scalp topography to audit alignment with functional neuroanatomy. \\ \midrule

\multicolumn{2}{@{}l@{}}{\textbf{3. Spectral Probing: Band-Wise Spectral Ablation (SA)}} \\
\hspace{0.3em}Masking & Given $\mathbf{Y} {\leftarrow} \mathrm{rFFT}(\mathbf{X})$ and band $[f_{low}, f_{high}]$, set binary mask $\mathbf{M}_b \in \{0,1\}^{1 \times F}$ to $M_b[k]{=}0$ if $f(k) \in [f_{low}, f_{high}]$. \\
\hspace{0.3em}Surrogate & Zero out target frequency band and reconstruct:\newline
$\hat{\mathbf{X}} {\leftarrow} \mathrm{irFFT}(\mathbf{Y} {\odot} \mathbf{M}_b,\allowbreak n{=}T)$. \\
\hspace{0.3em}Property & Isolates model reliance on physiological rhythms (e.g., $\delta, \theta, \alpha, \beta, \gamma$). \\ \bottomrule
\end{tabularx}
\end{table}

\begin{figure}[t]
\centering
\includegraphics[height=11cm ,width=22cm,angle=0,scale=0.4]{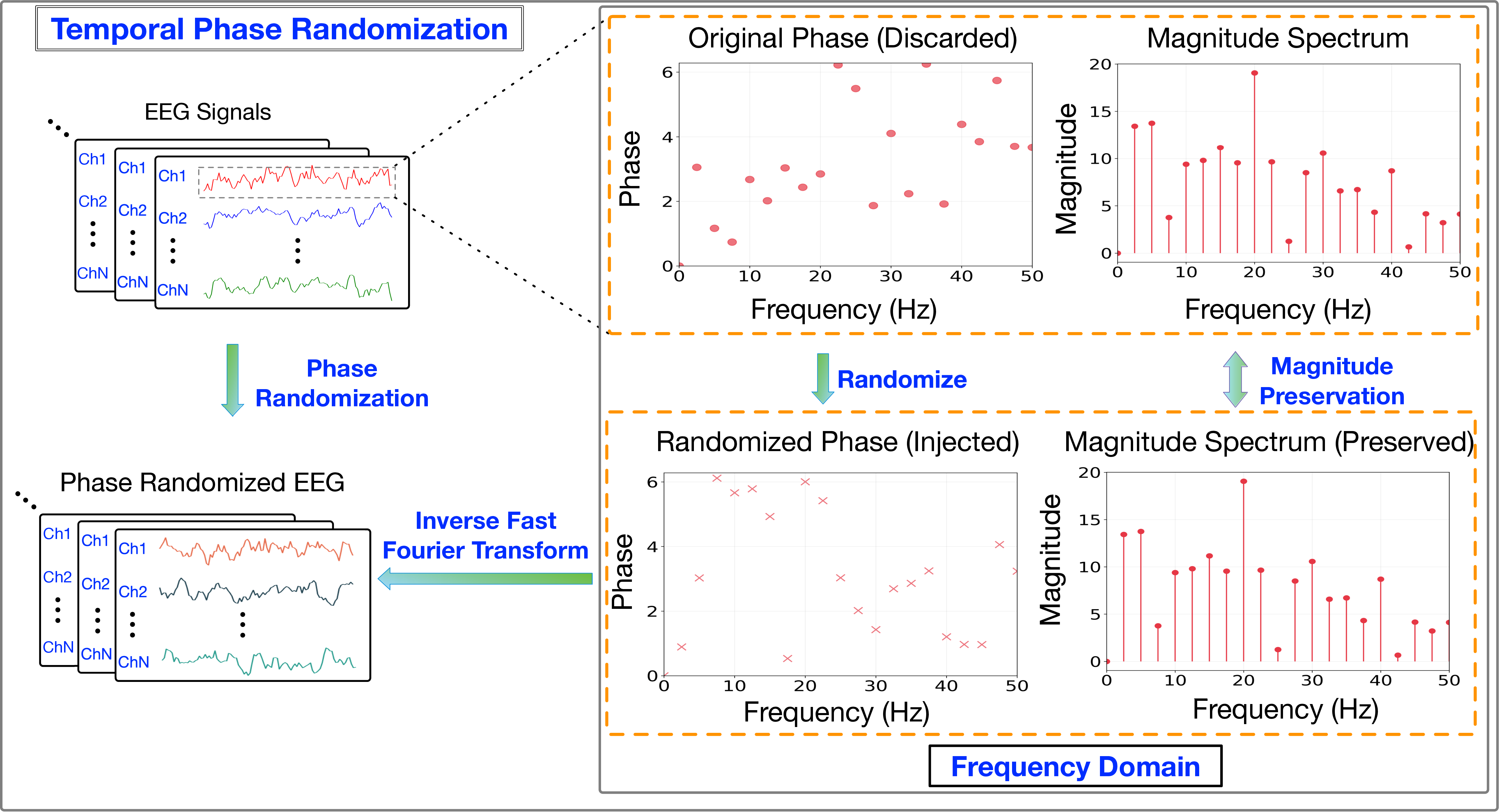}
\caption{Illustration of Fourier Phase Randomization. The original EEG signals are transformed into the frequency domain, where the original phase is replaced with a uniformly distributed, randomly generated phase $\phi \sim \mathcal{U}(0, 2\pi)$ to destroy phase information completely, while the magnitude is preserved. By applying identical phase randomization across all channels $C $, the inverse-transformed surrogate signals retain each channel's original frequency power spectrum and variance, but lose all temporal phase-locked structure.}
\label{fig_NPP_PR}
\end{figure}

Unlike traditional ``black-box'' evaluations, our proposed NPP framework (illustrated in Fig.~\ref{fig_NPP} and formalized in Table~\ref{tab:npp_modules}) systematically transforms an input EEG trial $\mathbf{X} \in \mathbb{R}^{C \times T}$ ($C$: channels, $T$: time points) into a controlled surrogate signal $\hat{\mathbf{X}} \in \mathbb{R}^{C \times T}$ across three fundamental dimensions to audit the neurophysiological feature dependencies of finetuned EEG-FMs.

\textbf{First, Temporal Probing} (see Fig.~\ref{fig_NPP_PR}) examines whether an EEG-FM genuinely relies on phase-locked temporal dynamics via Fourier phase randomization~\cite{prichard1994generating}. Given the zero-mean channel signal $\tilde{\mathbf{X}} = \mathbf{X} - \bar{\mathbf{X}}$, we apply the real Fast Fourier Transform ($\mathrm{rFFT}$) to obtain its complex spectrum $\mathbf{Y} = \mathrm{rFFT}(\tilde{\mathbf{X}}) = |\mathbf{Y}|e^{j\phi} \in \mathbb{C}^{C \times F}$ across $F = \lfloor T/2 \rfloor + 1$ frequency bins, where $|\mathbf{Y}|$ and $\phi$ denote the amplitude and phase, respectively. We then replace the original phases $\phi$ across all $C$ channels with a shared random phase vector $\theta \sim \mathcal{U}(0, 2\pi)^{1 \times F}$ across non-DC frequencies to reconstruct the temporal surrogate:
\begin{equation}
\hat{\mathbf{X}}_{\mathrm{temp}} = \mathrm{irFFT}\big(|\mathbf{Y}| \odot e^{j\theta},\, n=T\big) + \bar{\mathbf{X}},
\end{equation}
where $\mathrm{irFFT}(\cdot)$ denotes the inverse real Fast Fourier Transform and $\odot$ is the Hadamard product. This surrogate signal strictly preserves each channel's power spectrum while destroying phase-locked temporal structure.

\begin{figure}[t]
\centering
\includegraphics[height=11cm ,width=22cm,angle=0,scale=0.4]{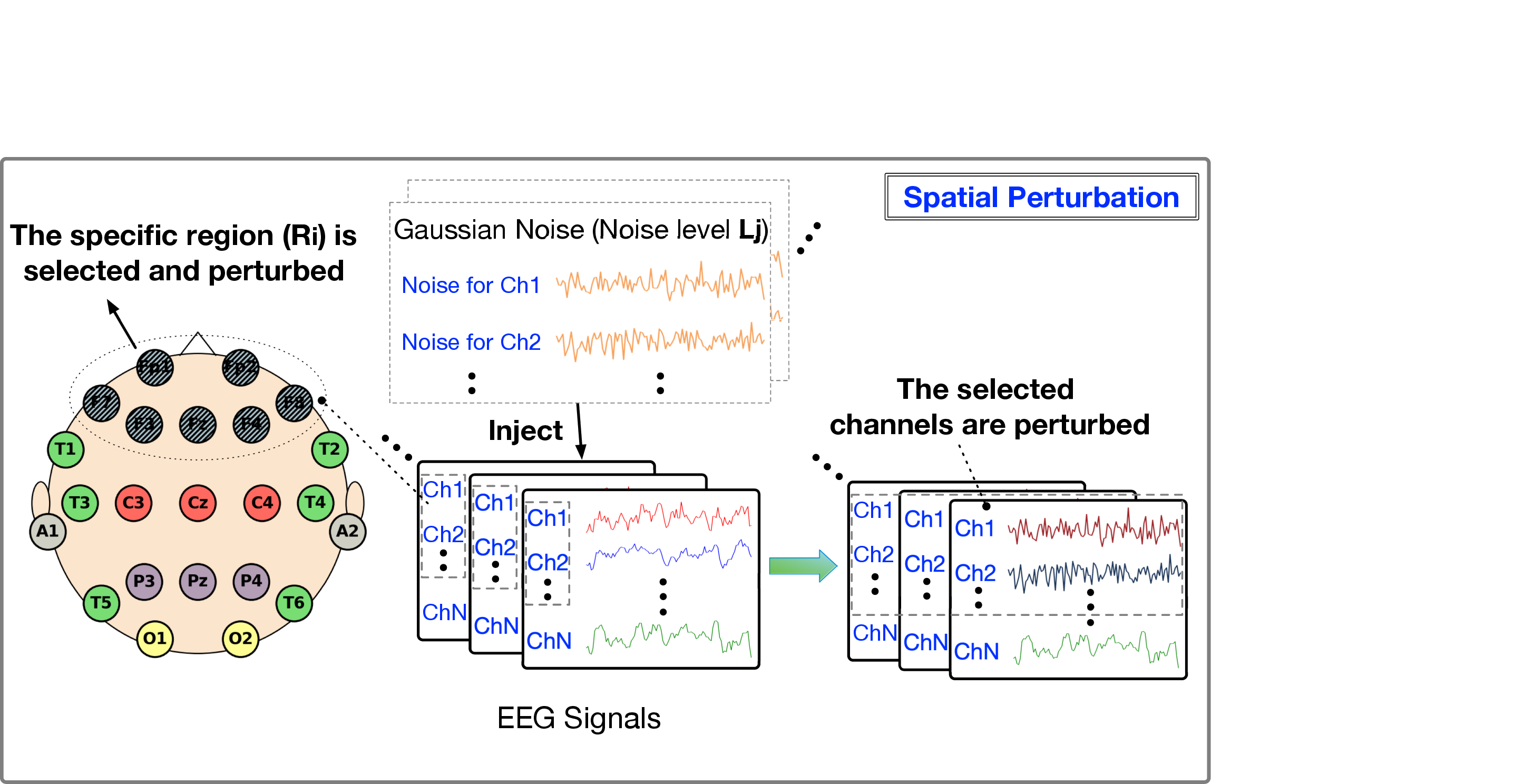}
\caption{Illustration of Spatial Perturbation. The given level $L_{j}$ of signal-aware Gaussian noises is respectively injected into the corresponding channels associated with the specific ROIs.}
\label{fig_NPP_SP}
\end{figure}

\textbf{Second, Spatial Probing} (see Fig.~\ref{fig_NPP_SP}) quantifies the contribution of specific brain regions to model predictions and evaluates alignment with functional neuroanatomy. Given a binary region-of-interest (ROI) channel mask $\mathbf{M}_r \in \{0, 1\}^{C \times 1}$ and global signal standard deviation $\sigma_{\mathbf{X}} = \mathrm{std}(\mathbf{X})$, we inject scale-aware Gaussian noise $\mathbf{Z} \sim \mathcal{N}(0, 1)^{C \times T}$ scaled by perturbation intensity $\lambda$ selectively into the target ROI channels:
\begin{equation}
\hat{\mathbf{X}}_{\mathrm{spat}} = \mathbf{X} + \lambda \cdot \sigma_{\mathbf{X}} \cdot (\mathbf{M}_r \odot \mathbf{Z}).
\end{equation}

\begin{figure}[t]
\centering
\includegraphics[height=11cm ,width=22cm,angle=0,scale=0.4]{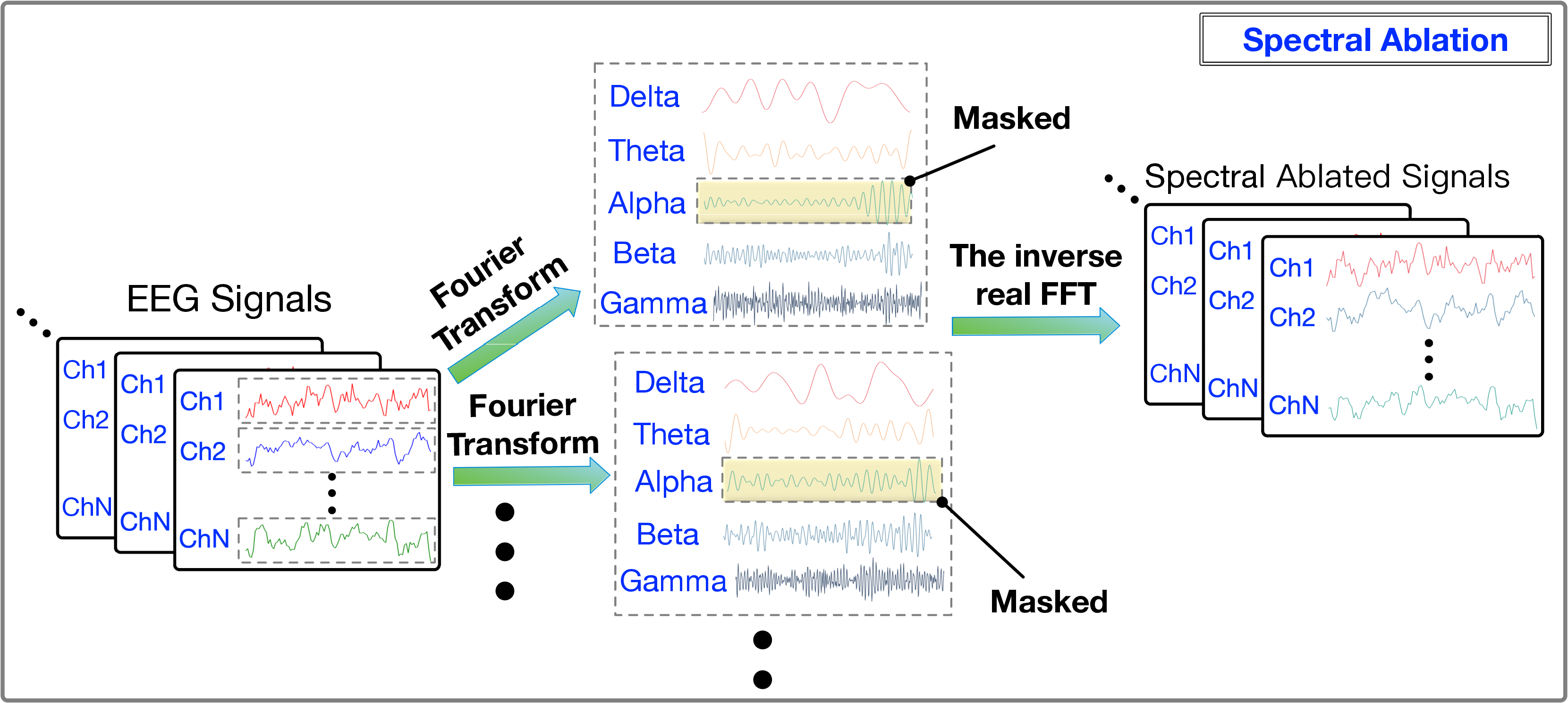}
\caption{Illustration of Spectral Ablation. The original EEG signals are first transformed into the frequency domain via rFFT, which decomposes the signals into physiological bands, e.g., Delta, Theta, Alpha, Beta and Gamma. Then, we ablate the selected frequency band by zeroing out its corresponding frequency coefficients. Finally, an inverse real FFT reconstructs the surrogate data back into the time domain. }
\label{fig_NPP_SM}
\end{figure}

\textbf{Finally, Spectral Probing} (see Fig.~\ref{fig_NPP_SM}) assesses whether an EEG-FM relies on frequency-specific features for its decision-making via band-wise spectral ablation. For a physiological rhythm $b \in \{\text{Delta, Theta, Alpha, Beta, Gamma}\}$ spanning $[f_{\mathrm{low}}, f_{\mathrm{high}}]$, we construct a binary frequency mask $\mathbf{M}_b \in \{0, 1\}^{1 \times F}$ where $M_b[k] = 0$ if frequency bin $f(k) \in [f_{\mathrm{low}}, f_{\mathrm{high}}]$ and $1$ otherwise, reconstructing the band-ablated signal via:
\begin{equation}
\hat{\mathbf{X}}_{\mathrm{spec}} = \mathrm{irFFT}\big(\mathrm{rFFT}(\mathbf{X}) \odot \mathbf{M}_b,\, n=T\big).
\end{equation}
By measuring the downstream performance degradation $\Delta$ of fine-tuned EEG-FMs induced by $\hat{\mathbf{X}}_{\mathrm{temp}}$, $\hat{\mathbf{X}}_{\mathrm{spat}}$, and $\hat{\mathbf{X}}_{\mathrm{spec}}$ (Fig.~\ref{fig_NPP} (b)), NPP provides a rigorous diagnostic benchmark to verify whether these models leverage neurophysiologically valid temporal, spatial, and spectral representations.

\section{Experiments}
\label{Experiment}

In this section, we detail the experimental setup, including the evaluated models, datasets, implementation details, and evaluation metrics. 


\textbf{Models.} We evaluated and analyzed four recent state-of-the-art EEG-FMs, namely NeuroGPT (Full Model)~\cite{cui2024neurogpt}, LaBraM-Base~\cite{jiang2024large}, NeuroLM-B~\cite{jiang2025neurolm}, and EEGPT-Large~\cite{wang2024eegpt}, as well as representative traditional supervised models, including EEGNet~\cite{Lawhern2018EEGNet}, CSPNet~\cite{Jiang2024CSPNet}, TS-SEFFNet~\cite{Li2021TSSEFFNet}, MSCFormer~\cite{Zhao2025MSCFormer}, and CTNet~\cite{Zhao2024CTNet}. Hereafter, we refer to the evaluated FMs simply as NeuroGPT, LaBraM, NeuroLM, and EEGPT, respectively.

\textbf{Datasets.} In our experiments, we use four EEG datasets: TUAB~\cite{lopez2015automated}, TUEV~\cite{harati2015improved}, BCI Competition IV-2b~\cite{leeb2008bci}, and SEED-IV~\cite{zheng2018emotionmeter}. TUAB is a large-scale clinical dataset designed for binary classification (abnormal vs. normal). This dataset contains a total of 409,445 10-second EEG trials with 23 channels at 256 Hz. The TUEV dataset targets multi-class EEG event detection and contains six annotated event types, including epileptiform events (spikes and sharp waves, generalized periodic epileptiform discharges, and periodic lateralized epileptiform discharges) and non-epileptiform events (artifacts, eye movements, and background activity). It comprises 112,491 five-second trials recorded from 23 channels at 256 Hz. For TUAB and TUEV, we use the official LaBraM preprocessing scripts\footnote{\url{https://github.com/935963004/LaBraM}}, applying $0.1\text{--}75\text{ Hz}$ bandpass filtering, $50\text{ Hz}$ notch filtering, and resampling to $200\text{ Hz}$. BCI Competition IV-2b is a widely used benchmark for two-class motor imagery (left vs. right hand) BCI classification. It includes recordings from nine subjects across five sessions (two non-feedback screening sessions $\text{01T}$–$\text{02T}$ and three online smiley-feedback sessions $\text{03T}$, $\text{04E}$, and $\text{05E}$), acquired using three bipolar EEG channels ($\text{C3}$, $\text{Cz}$, $\text{C4}$) and three EOG channels at $250\text{ Hz}$ with hardware-based bandpass filtering (0.5–100 Hz) and 50 Hz notch filtering. In our experiments, we use only the non-feedback sessions ($\text{01T}$–$\text{02T}$, 240–280 trials per subject) to preserve paradigm homogeneity and avoid visual-feedback confounds. We retain the three bipolar EEG channels and extract a $2.0\text{ s}$ motor imagery window per trial. For simplicity, we refer to BCI Competition IV-2b as BCI IV 2b. SEED-IV dataset is an established affective EEG dataset designed for four-class emotion recognition (neutral, sad, fear, and happy). It contains recordings from 15 subjects across three sessions (24 trials per session) elicited by movie clips, which is collected using a 62-channel montage, bandpass-filtered ($1\text{–}75\text{ Hz}$), and downsampled to $200\text{ Hz}$. In our experiments, each variable-length trial is segmented into complete, non-overlapping $4.0\text{ s}$ windows. Note that across all datasets, we make adjustments to meet specific model requirements based on their original papers (e.g., resampling to $200\text{ Hz}$ for LaBraM).

\textbf{Experimental Details.} For the two large-scale datasets (TUAB and TUEV), we follow the data splitting strategy used in representative works~\cite{jiang2024large}, dividing each dataset into training, validation, and test sets. To ensure robustness, each model is trained and evaluated five times using fixed random seeds, and the results are averaged across runs. We monitor balanced accuracy on validation data to decide the best model and then test the best model on test data. For the BCI Competition IV-2b dataset, due to its limited size (9 subjects), we adopt a leave-one-subject-out (LOSO) cross-validation protocol. In each fold, data from 8 subjects are used for model development, with internal validation splits applied where required by the hyperparameter tuning strategy, while the remaining unseen subject is held out exclusively for testing. This process is repeated until all 9 subjects have been evaluated. We adopt an identical LOSO evaluation protocol and fixed-epoch training strategy for SEED-IV across its 15 subjects. In our experiments, we use the publicly available source code for the evaluated models. All experiments are conducted on two servers, equipped with 4 NVIDIA A100 GPUs and 8 NVIDIA A5000 GPUs, respectively.

\textbf{Evaluation Metrics.} To ensure a robust evaluation across both binary and multi-class EEG decoding tasks, we adopt three complementary metrics: Balanced Accuracy (BA), Macro F1-Score (F1), and Cohen's Kappa ($\kappa$). Specifically, BA calculates the average recall across all classes, giving each class equal weight and preventing majority classes from dominating the results in imbalanced clinical datasets (e.g., TUEV). Macro-F1 calculates the average F1-score across all classes, while $\kappa$ measures the agreement between model predictions and ground-truth labels, adjusted for chance.

\begin{table*}[t]
\centering
\fontsize{7pt}{8.8pt}\selectfont
\setlength{\tabcolsep}{3.5pt}
\medmuskip=1mu
\renewcommand{\arraystretch}{1.15}
\caption{Comprehensive performance comparison of Balanced Accuracy (BA), Macro F1-Score (F1), and Cohen's Kappa ($\kappa$) across TUEV, TUAB, and BCI-IV-2b (see Table~\ref{tab:seed_iv_performance} for SEED-IV). EEG-FMs are denoted in \textit{italics}. The $\dagger$ symbol represents default configurations, while $*$ denotes our fairness-calibrated configurations. \protect\underline{Underlined values} indicate instances where the optimized configuration outperformed default settings. Results are reported as Mean $\mu \pm$ Standard Deviation $\sigma$. The \first{1\textsuperscript{st}}, \second{2\textsuperscript{nd}}, and \third{3\textsuperscript{rd}} results are highlighted. The hyperparameter search spaces and selected configurations for both supervised baselines and EEG-FMs across all four datasets are detailed in the Supplementary Material.}
\label{ASHABench}
\begin{tabular}{l ccc ccc ccc}
\toprule
\multirow{2.5}{*}{\textbf{Model}} & \multicolumn{3}{c}{\textbf{TUEV}} & \multicolumn{3}{c}{\textbf{TUAB}} & \multicolumn{3}{c}{\textbf{BCI-IV-2b}} \\
\cmidrule(lr){2-4} \cmidrule(lr){5-7} \cmidrule(lr){8-10}
& BA & F1 & $\kappa$ & BA & F1 & $\kappa$ & BA & F1 & $\kappa$ \\
\midrule
EEGNet$\dagger$      & $0.221 \pm 0.055$ & $0.229 \pm 0.080$ & $0.098 \pm 0.097$ & $0.719 \pm 0.059$ & $0.707 \pm 0.074$ & $0.450 \pm 0.120$ & $0.498 \pm 0.040$ & $0.493 \pm 0.043$ & $-0.004 \pm 0.080$ \\
CSPNet$\dagger$      & $0.395 \pm 0.074$ & $0.339 \pm 0.041$ & $0.408 \pm 0.083$ & $0.705 \pm 0.029$ & $0.698 \pm 0.038$ & $0.407 \pm 0.061$ & $0.625 \pm 0.074$ & $0.615 \pm 0.077$ & $0.250 \pm 0.148$ \\
TS-SEFFNet$\dagger$  & $0.418 \pm 0.007$ & $0.429 \pm 0.009$ & $0.360 \pm 0.019$ & \third{$0.800 \pm 0.003$} & \first{$0.802 \pm 0.003$} & \third{$0.605 \pm 0.008$} & $0.598 \pm 0.048$ & $0.593 \pm 0.050$ & $0.197 \pm 0.098$ \\
MSCFormer$\dagger$   & $0.426 \pm 0.012$ & $0.417 \pm 0.012$ & $0.353 \pm 0.043$ & $0.733 \pm 0.026$ & $0.733 \pm 0.030$ & $0.473 \pm 0.052$ & \second{$0.658 \pm 0.085$} & \second{$0.651 \pm 0.089$} & \second{$0.317 \pm 0.170$} \\
CTNet$\dagger$       & $0.207 \pm 0.028$ & $0.210 \pm 0.047$ & $0.091 \pm 0.062$ & $0.790 \pm 0.015$ & $0.792 \pm 0.015$ & $0.585 \pm 0.029$ & $0.599 \pm 0.072$ & $0.589 \pm 0.079$ & $0.199 \pm 0.144$ \\
\midrule
EEGNet$*$            & \underline{$0.261 \pm 0.041$} & \underline{$0.284 \pm 0.053$} & \underline{$0.284 \pm 0.072$} & \underline{$0.781 \pm 0.008$} & \underline{$0.783 \pm 0.009$} & \underline{$0.571 \pm 0.016$} & \underline{\first{$0.680 \pm 0.049$}} & \underline{\first{$0.669 \pm 0.050$}} & \underline{\first{$0.359 \pm 0.099$}} \\
CSPNet$*$            & $0.386 \pm 0.078$ & $0.336 \pm 0.074$ & $0.338 \pm 0.099$ & \underline{$0.779 \pm 0.005$} & \underline{$0.781 \pm 0.005$} & \underline{$0.565 \pm 0.011$} & $0.603 \pm 0.059$ & $0.597 \pm 0.061$ & $0.207 \pm 0.118$ \\
TS-SEFFNet$*$        & \underline{$0.486 \pm 0.009$} & \underline{\third{$0.496 \pm 0.009$}} & \underline{$0.467 \pm 0.028$} & $0.796 \pm 0.004$ & \second{$0.798 \pm 0.004$} & $0.598 \pm 0.006$ & \underline{$0.630 \pm 0.070$} & \underline{$0.616 \pm 0.078$} & \underline{$0.261 \pm 0.133$} \\
MSCFormer$*$         & $0.340 \pm 0.039$ & $0.343 \pm 0.035$ & \underline{$0.388 \pm 0.067$} & \underline{$0.797 \pm 0.005$} & \underline{\third{$0.797 \pm 0.005$}} & \underline{$0.594 \pm 0.011$} & $0.639 \pm 0.058$ & $0.628 \pm 0.061$ & $0.278 \pm 0.115$ \\
CTNet$*$             & \underline{$0.271 \pm 0.021$} & \underline{$0.271 \pm 0.021$} & \underline{$0.193 \pm 0.009$} & $0.784 \pm 0.010$ & $0.785 \pm 0.010$ & $0.575 \pm 0.018$ & \underline{$0.639 \pm 0.083$} & \underline{$0.630 \pm 0.091$} & \underline{$0.279 \pm 0.166$} \\
\midrule
\textit{NeuroGPT}$\dagger$ & $0.357 \pm 0.016$ & $0.351 \pm 0.012$ & $0.360 \pm 0.020$ & $0.777 \pm 0.002$ & $0.781 \pm 0.001$ & $0.558 \pm 0.003$ & \third{$0.645 \pm 0.069$} & \third{$0.645 \pm 0.065$} & \third{$0.290 \pm 0.130$} \\
\textit{LaBraM}$\dagger$   & $0.511 \pm 0.015$ & $0.433 \pm 0.013$ & $0.402 \pm 0.012$ & \second{$0.803 \pm 0.007$} & $0.782 \pm 0.008$ & \second{$0.610 \pm 0.015$} & $0.532 \pm 0.013$ & $0.511 \pm 0.060$ & $0.064 \pm 0.026$ \\
\textit{NeuroLM}$\dagger$  & $0.468 \pm 0.018$ & $0.386 \pm 0.025$ & $0.371 \pm 0.017$ & $0.791 \pm 0.002$ & $0.776 \pm 0.005$ & $0.584 \pm 0.006$ & $0.586 \pm 0.070$ & $0.571 \pm 0.070$ & $0.172 \pm 0.138$ \\
\textit{EEGPT}$\dagger$    & \third{$0.537 \pm 0.019$} & \second{$0.510 \pm 0.023$} & \second{$0.501 \pm 0.021$} & $0.792 \pm 0.003$ & $0.794 \pm 0.003$ & $0.589 \pm 0.007$ & $0.577 \pm 0.095$ & $0.546 \pm 0.135$ & $0.154 \pm 0.190$ \\
\midrule
\textit{NeuroGPT}$*$ & \underline{$0.392 \pm 0.028$} & \underline{$0.370 \pm 0.026$} & \underline{$0.386 \pm 0.022$} & \underline{$0.778 \pm 0.003$} & \underline{$0.782 \pm 0.002$} & \underline{$0.560 \pm 0.004$} & $0.602 \pm 0.059$ & $0.599 \pm 0.058$ & $0.205 \pm 0.117$ \\
\textit{LaBraM}$*$   & \underline{\second{$0.547 \pm 0.020$}} & \underline{$0.488 \pm 0.028$} & \underline{\third{$0.493 \pm 0.038$}} & \underline{\first{$0.808 \pm 0.003$}} & \underline{\third{$0.797 \pm 0.004$}} & \underline{\first{$0.630 \pm 0.007$}} & \underline{$0.544 \pm 0.019$} & \underline{$0.543 \pm 0.019$} & \underline{$0.088 \pm 0.038$} \\
\textit{NeuroLM}$*$  & \underline{$0.494 \pm 0.016$} & $0.364 \pm 0.028$ & $0.335 \pm 0.035$ & $0.791 \pm 0.003$ & $0.773 \pm 0.005$ & $0.581 \pm 0.003$ & \underline{$0.610 \pm 0.074$} & \underline{$0.603 \pm 0.077$} & \underline{$0.218 \pm 0.147$} \\
\textit{EEGPT}$*$    & \underline{\first{$0.562 \pm 0.038$}} & \underline{\first{$0.518 \pm 0.034$}} & \underline{\first{$0.535 \pm 0.035$}} & $0.788 \pm 0.004$ & $0.766 \pm 0.004$ & $0.580 \pm 0.009$ & $0.572 \pm 0.038$ & $0.536 \pm 0.040$ & $0.144 \pm 0.076$ \\
\bottomrule
\end{tabular}
\end{table*}

\begin{table}[t]
\centering
\fontsize{9pt}{8.8pt}\selectfont
\setlength{\tabcolsep}{4.5pt}
\medmuskip=1mu
\renewcommand{\arraystretch}{1.15}
\caption{Performance comparison of Balanced Accuracy (BA), Macro F1-Score (F1), and Cohen's Kappa ($\kappa$) on SEED-IV. Notation follows Table~\ref{ASHABench}.}
\label{tab:seed_iv_performance}
\begin{tabular}{l ccc}
\toprule
\textbf{Model} & \textbf{BA} & \textbf{F1} & $\boldsymbol{\kappa}$ \\
\midrule
EEGNet$\dagger$            & $0.339 \pm 0.050$ & $0.304 \pm 0.059$ & $0.115 \pm 0.065$ \\
CSPNet$\dagger$            & $0.264 \pm 0.021$ & $0.152 \pm 0.049$ & $0.017 \pm 0.027$ \\
TS-SEFFNet$\dagger$        & $0.322 \pm 0.057$ & $0.271 \pm 0.083$ & $0.093 \pm 0.073$ \\
MSCFormer$\dagger$         & $0.255 \pm 0.010$ & $0.124 \pm 0.030$ & $0.007 \pm 0.015$ \\
CTNet$\dagger$             & \third{$0.363 \pm 0.045$} & \third{$0.336 \pm 0.057$} & \second{$0.150 \pm 0.057$} \\
\midrule
EEGNet$*$                  & $0.302 \pm 0.053$ & $0.234 \pm 0.080$ & $0.073 \pm 0.073$ \\
CSPNet$*$                  & \underline{\second{$0.369 \pm 0.061$}} & \underline{\second{$0.341 \pm 0.068$}} & \underline{$0.159 \pm 0.081$} \\
TS-SEFFNet$*$              & $0.322 \pm 0.048$ & $0.263 \pm 0.078$ & $0.093 \pm 0.061$ \\
MSCFormer$*$               & \underline{$0.352 \pm 0.052$} & \underline{$0.293 \pm 0.072$} & \underline{$0.135 \pm 0.068$} \\
CTNet$*$                   & $0.356 \pm 0.066$ & $0.322 \pm 0.085$ & \third{$0.140 \pm 0.085$} \\
\midrule
\textit{NeuroGPT}$\dagger$ & $0.306 \pm 0.032$ & $0.294 \pm 0.032$ & $0.073 \pm 0.040$ \\
\textit{LaBraM}$\dagger$   & $0.308 \pm 0.031$ & $0.241 \pm 0.045$ & $0.075 \pm 0.040$ \\
\textit{NeuroLM}$\dagger$  & $0.305 \pm 0.040$ & $0.277 \pm 0.051$ & $0.070 \pm 0.050$ \\
\textit{EEGPT}$\dagger$    & $0.308 \pm 0.039$ & $0.272 \pm 0.056$ & $0.080 \pm 0.054$ \\
\midrule
\textit{NeuroGPT}$*$       & \underline{$0.341 \pm 0.067$} & \underline{$0.312 \pm 0.066$} & \underline{$0.120 \pm 0.087$} \\
\textit{LaBraM}$*$         & $0.284 \pm 0.025$ & \underline{$0.270 \pm 0.030$} & $0.045 \pm 0.033$ \\
\textit{NeuroLM}$*$        & \underline{$0.324 \pm 0.036$} & \underline{$0.305 \pm 0.038$} & \underline{$0.095 \pm 0.047$} \\
\textit{EEGPT}$*$          & \underline{\first{$0.389 \pm 0.050$}} & \underline{\first{$0.366 \pm 0.051$}} & \underline{\first{$0.181 \pm 0.066$}} \\
\bottomrule
\end{tabular}
\end{table}

\section{Results and Discussion}

In this section, we present extensive experimental results and key findings obtained with the proposed EEG-FM-Audit framework. First, Section~\ref{FMBenchmarking} presents the benchmarking results for both default and well-tuned models across three metrics and four datasets. Next, we report our paradigm-level ablation results in Section~\ref{Exp_3PLAB} and our neurophysiological probing (NPP) findings in Section~\ref{Exp_NPPR}. To further validate the reliability of NPP, we conduct spatial region-of-interest (ROI) perturbation and equal-width spectral ablation experiments in Section~\ref{app:npp_control}. Finally, we discuss the rationale behind our asymmetric optimization strategy (Fairness-Calibrated Benchmarking) in Section~\ref{app:fm_AOS}.

\subsection{Benchmarking ASHA-Optimized Baselines against Tuned EEG Foundation Models} 
\label{FMBenchmarking}

We evaluate optimized EEG-FMs against carefully tuned supervised baselines to provide a fair comparison. Based on the experimental results, we have several key observations.

\begin{observationbox}{Observation 1.}
While foundation models show consistently superior performance on complex, large-scale clinical datasets like TUEV, well-tuned supervised baselines remain highly competitive. Specifically, on tasks such as motor imagery (BCI IV 2b), the well-tuned baselines can match or even outperform the well-tuned EEG-FMs, though they have much fewer parameters.
\end{observationbox}

\begin{figure}[t]
\centering
\includegraphics[height=13cm ,width=22cm,angle=0,scale=0.4]{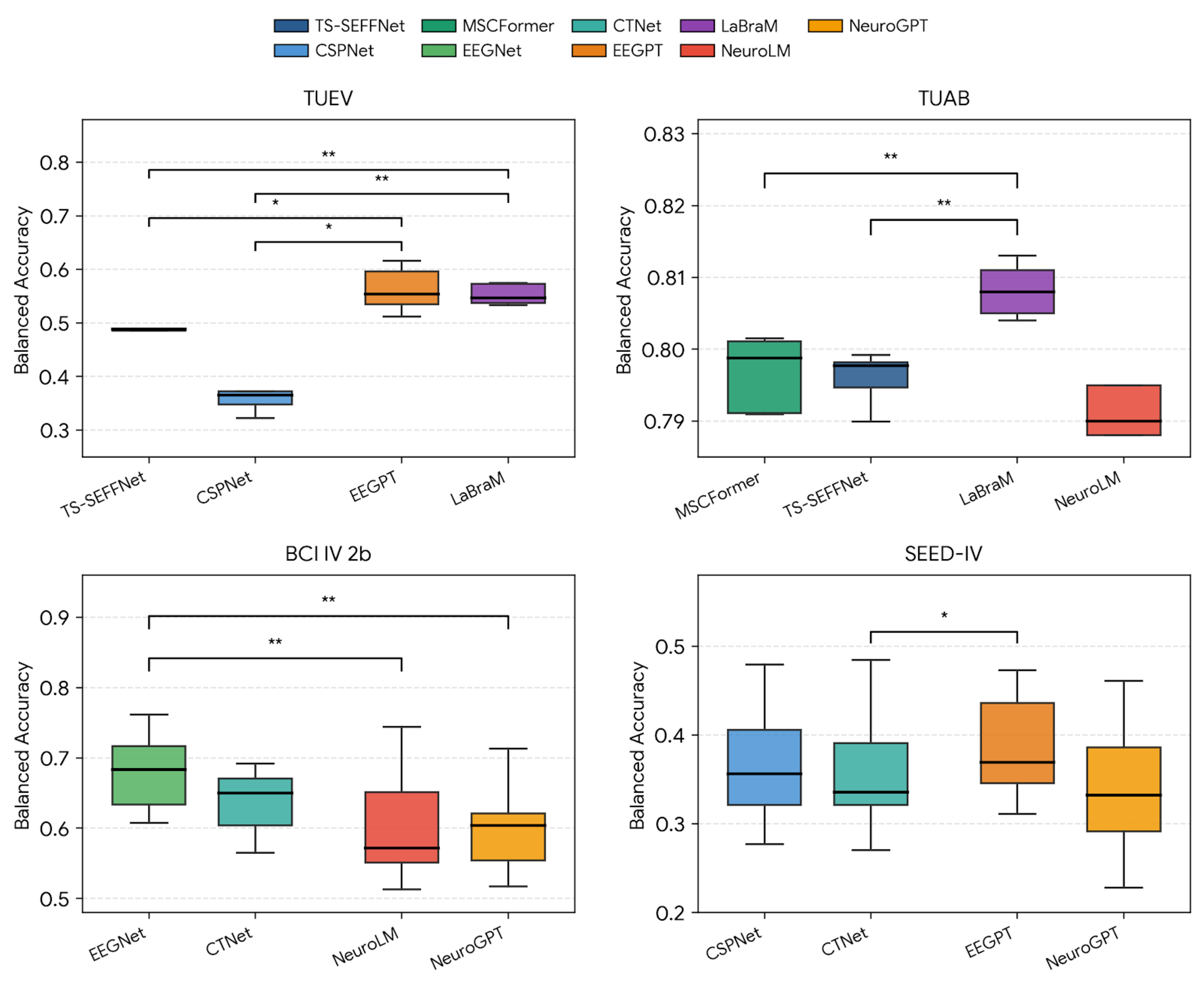}
\caption{Significance analysis of top-performing models across datasets. Horizontal brackets denote two-tailed paired $t$-tests evaluated across matched evaluation runs ($^*p < 0.05$, $^{**}p < 0.01$). On large downstream datasets (TUAB and TUEV), leading EEG-FMs significantly outperform top-performing supervised models. Conversely, on smaller datasets, the foundation model advantage diminishes. For example, on BCI IV 2b, the supervised baseline EEGNet significantly outperforms top-performing EEG-FMs ($p < 0.01$), whereas on SEED-IV, the foundation model EEGPT significantly outperforms one of the leading supervised models, CTNet ($p < 0.05$) while remaining comparable to CSPNet.}
\label{fig_SAModels}
\end{figure}

As shown in Table~\ref{ASHABench} and~\ref{tab:seed_iv_performance}, the top-performing EEG-FMs consistently rank among the top three across all four datasets and metrics, although they require substantially larger parameter counts (see Table~\ref{tab:model_params}). In particular, on large-scale downstream datasets (TUEV and TUAB), leading EEG-FMs such as LaBraM significantly outperform the best supervised baselines ($p < 0.05$ or $p < 0.01$; see Fig.~\ref{fig_SAModels}). This demonstrates the strong ability of representative EEG-FMs to perform diverse EEG decoding tasks. However, well-tuned supervised models can match or even outperform leading EEG-FMs on the small downstream datasets. For example, a simple supervised baseline (EEGNet) achieves significantly better performance than leading EEG-FMs on BCI IV 2b ($p < 0.01$; see Fig.~\ref{fig_SAModels}), while CSPNet matches the performance of the best-performing EEG-FMs on SEED-IV (no significant difference; see Fig.~\ref{fig_SAModels}).

\begin{table}[ht]
\centering
\scriptsize
\setlength{\tabcolsep}{8pt}
\renewcommand{\arraystretch}{1.12}
\caption{Model parameter counts across evaluated architectures. Supervised baseline parameter counts vary across the four datasets due to different architectural hyperparameters selected from the optimization search space (as well as dataset-specific input dimensions), whereas EEG-FMs ({\normalfont\itshape italicized}) maintain fixed pretrained backbone sizes.}
\label{tab:model_params}
\begin{tabular}{llc}
\toprule
\textbf{Category} & \textbf{Model} & \textbf{Parameters} \\
\midrule
\multirow{5}{*}{Supervised Baselines} 
 & EEGNet     & $\approx$ 1.6k -- 9.4k \\
 & CSPNet     & $\approx$ 0.7k -- 25k \\
 & TS-SEFFNet & $\approx$ 321k -- 337k \\
 & MSCFormer  & $\approx$ 26k -- 1.1M \\
 & CTNet      & $\approx$ 71k -- 150k \\
\midrule
\multirow{4}{*}{EEG Foundation Models} 
 & \textit{NeuroGPT} & $\approx$ 79.8M \\
 & \textit{LaBraM}   & $\approx$ 5.8M \\
 & \textit{NeuroLM}  & $\approx$ 253M \\
 & \textit{EEGPT}    & $\approx$ 101M \\
\bottomrule
\end{tabular}
\end{table}

\begin{figure}[t]
\centering
\includegraphics[width=\columnwidth]{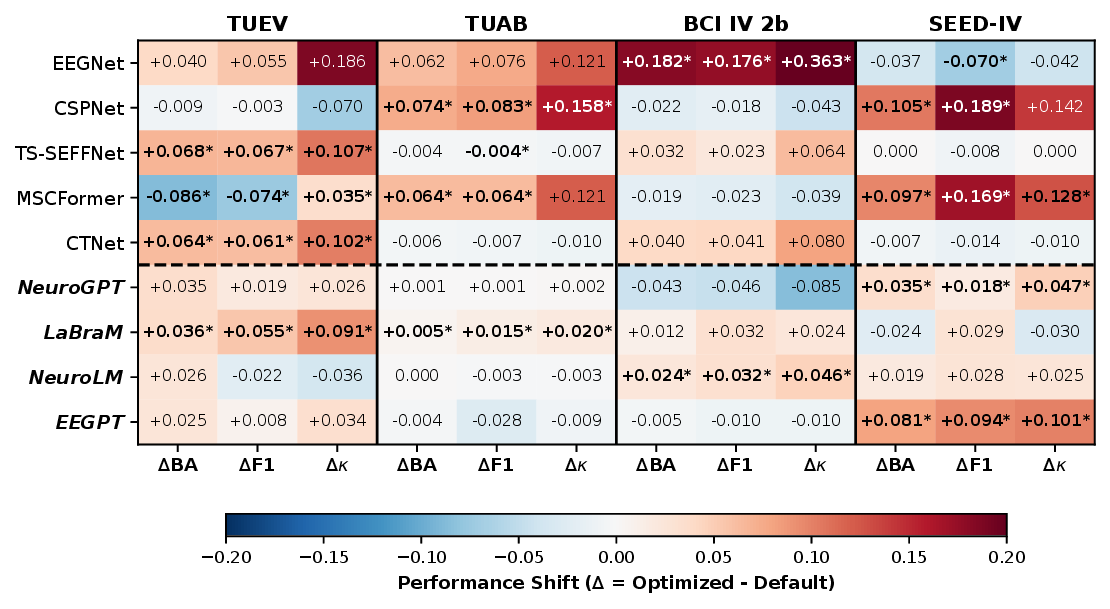}
\caption{Hyperparameter Optimization Effectiveness across four datasets. $\Delta$ denotes the performance gain from using fairness-calibrated (optimized) hyperparameters compared to default settings ($\Delta = \text{Optimized}^* - \text{Default}^\dagger$). The horizontal dashed line separates supervised baselines (top) from pretrained FMs (bottom, italicized). Statistically significant differences compared to the default setting are bolded and marked with an asterisk ($*$) (two-tailed paired $t$-test across seeds/folds, $p < 0.05$).}
\label{Fig_ASHA_Effectiveness}
\end{figure}

\begin{observationbox}{Observation 2.} Systematic HPO yields performance improvements in most cases across all four datasets for both supervised models and EEG-FMs. In particular, the gains for supervised baselines are often substantial, underscoring the critical need for rigorous hyperparameter tuning when benchmarking foundation models against traditional approaches.
\end{observationbox}

While HPO occasionally leads to performance decreases due to ASHA early stopping or finite hyperparameter search spaces, the majority of these decreases are not significant ($p > 0.05$; see Fig.~\ref{Fig_ASHA_Effectiveness}). HPO improves performance across most metrics on all evaluated datasets over default baselines, with significant Balanced Accuracy improvements in many cases (see Fig.~\ref{Fig_ASHA_Effectiveness}). Especially for supervised models, HPO leads to statistically significant performance improvements for most metrics on datasets such as TUAB and TUEV. Since existing studies~\cite{cui2024neurogpt} often omit or obscure tuning strategies when comparing foundation models to supervised baselines, they risk underestimating baseline performance and leading to unfair comparisons.

\subsection{Paradigm-Level Ablation Results}
\label{Exp_3PLAB}

To explore the effectiveness of key learning paradigms in EEG-FMs, we conduct three paradigm-level ablation studies: Pretraining-source ablation, LLM-Rep ablation and No-GPT ablation.  

\begin{observationbox}{Observation 3 on Pretraining-Source Ablation.} The effectiveness of large-scale pretraining depends on both model architecture and dataset scale. EEG-FMs based on discrete tokenization (e.g., LaBraM and NeuroLM) rely heavily on large-scale pretraining, whereas continuous-representation models (e.g., NeuroGPT and EEGPT) generally achieve competitive performance using target-only pretraining on large target datasets (e.g., TUAB and TUEV). However, across all architectures, target-only pretraining leads to degraded performance on small downstream datasets (e.g., BCI IV 2b and SEED-IV), suggesting that sufficient data scale remains critical for effective foundation model pretraining.   

\end{observationbox}

\begin{table}[ht]
\centering
\scriptsize 
\setlength{\tabcolsep}{2.5pt} 
\renewcommand{\arraystretch}{1.15}
\caption{Performance gains from large-scale heterogeneous pretraining across all four benchmarks. $\Delta$ denotes the performance difference between foundation models using large-scale pretraining and variants retrained only on target-domain data (Original $-$ Target-only), where positive values indicate gains from large-scale pretraining.}
\label{Tab_PSAB}

\begin{tabular}{l ccc ccc}
\toprule
\multirow{2.5}{*}{\textbf{Model}} & \multicolumn{3}{c}{\textbf{TUEV}} & \multicolumn{3}{c}{\textbf{TUAB}} \\
\cmidrule(lr){2-4} \cmidrule(lr){5-7}
& $\Delta$BA & $\Delta$F1 & $\Delta$Kappa & $\Delta$BA & $\Delta$F1 & $\Delta$Kappa \\[0.5ex]
\midrule
NeuroGPT & -0.033 \downgreen & -0.056 \downgreen & -0.036 \downgreen & +0.001 \upred     & +0.002 \upred & +0.004 \upred     \\
LaBraM   & +0.152 \upred     & +0.029 \upred     & +0.052 \upred     & +0.041 \upred     & +0.044 \upred & +0.085 \upred     \\
NeuroLM  & +0.102 \upred     & +0.083 \upred     & +0.107 \upred     & +0.031 \upred     & +0.044 \upred & +0.061 \upred     \\
EEGPT    & -0.001 \downgreen & -0.020 \downgreen & -0.113 \downgreen & -0.006 \downgreen & +0.018 \upred & -0.011 \downgreen \\
\bottomrule
\end{tabular}

\vspace{0.65em}

\begin{tabular}{l ccc ccc}
\toprule
\multirow{2.5}{*}{\textbf{Model}} & \multicolumn{3}{c}{\textbf{BCI IV 2b}} & \multicolumn{3}{c}{\textbf{SEED-IV}} \\
\cmidrule(lr){2-4} \cmidrule(lr){5-7}
& $\Delta$BA & $\Delta$F1 & $\Delta$Kappa & $\Delta$BA & $\Delta$F1 & $\Delta$Kappa \\[0.5ex]
\midrule
NeuroGPT & +0.094 \upred & +0.088 \upred & +0.188 \upred & +0.091 \upred & +0.205 \upred & +0.120 \upred \\
LaBraM   & +0.016 \upred & +0.009 \upred & +0.033 \upred & +0.029 \upred & +0.141 \upred & +0.038 \upred \\
NeuroLM  & +0.084 \upred & +0.101 \upred & +0.167 \upred & +0.042 \upred & +0.036 \upred & +0.053 \upred \\
EEGPT    & +0.049 \upred & +0.049 \upred & +0.098 \upred & +0.118 \upred & +0.170 \upred & +0.150 \upred \\
\bottomrule
\end{tabular}
\end{table}


\textbf{Pretraining-Source Ablation.}  As shown in Table~\ref{Tab_PSAB}, pretraining only on target datasets (TUAB and TUEV) generally improves or maintains the performance for continuous-representation models (NeuroGPT and EEGPT), where their encoder part directly maps EEG signals into a continuous vector space. In contrast, LaBraM and NeuroLM consistently exhibit performance degradation across all metrics on both datasets when pretrained solely on the target dataset, showing absolute drops of at least $0.031$, $0.029$, and $0.052$ in Balanced Accuracy, F1 score, and Kappa, respectively. A possible explanation for this discrepancy is that LaBraM and NeuroLM use discrete tokenization, which needs diverse data to learn a sufficiently expressive codebook. In contrast, NeuroGPT and EEGPT directly model the target distribution’s temporal and spectral dynamics. Furthermore, all models show consistent performance degradation on BCI IV-2b and SEED-IV under target-only pretraining. This degradation is likely because the limited dataset size is insufficient to support effective representation learning for large-scale EEG-FMs.

\begin{observationbox}{Observation 4 on LLM-Rep Ablation.} When the target dataset is small (e.g., BCI IV 2b), FMs almost consistently outperform their ablated counterparts. This highlights the necessity of the pretraining step when deploying FMs in low-data regimes. Similarly, performance often decreases on large datasets (e.g., TUEV) after removing the entire pretraining paradigm in FMs. 

\end{observationbox}

\textbf{LLM-Rep Ablation.} As shown in Table~\ref{Tab_LIAB}, on BCI IV 2b, all ablated models exhibit consistent performance drops of at least 0.017 across all three metrics. This is likely because limited supervision in a small dataset makes models highly sensitive to initialization. A similar trend appears on SEED-IV, with the exception of LaBraM's F1 score. For two larger target datasets, ablated models generally show performance degradation, although performance is occasionally maintained or even improved. For example, NeuroLM achieves modest gains of 0.01–0.02 in F1 score and Kappa on TUEV.

\begin{table}[ht]
\centering
\scriptsize 
\setlength{\tabcolsep}{2.5pt} 
\renewcommand{\arraystretch}{1.15}
\caption{Performance gains from the entire pretraining stages. $\Delta$ denotes the performance difference between FMs and their supervised counterparts (Original $-$ Ablated), where positive values indicate gains from pretraining.}
\label{Tab_LIAB}

\begin{tabular}{l ccc ccc}
\toprule
\multirow{2.5}{*}{\textbf{Model}} & \multicolumn{3}{c}{\textbf{TUEV}} & \multicolumn{3}{c}{\textbf{TUAB}} \\
\cmidrule(lr){2-4} \cmidrule(lr){5-7}
& $\Delta$BA & $\Delta$F1 & $\Delta$Kappa & $\Delta$BA & $\Delta$F1 & $\Delta$Kappa \\[0.5ex]
\midrule
NeuroGPT & +0.188 \upred & +0.329 \upred     & +0.363 \upred     & +0.006 \upred     & +0.006 \upred     & -0.002 \downgreen \\
LaBraM   & +0.047 \upred & +0.066 \upred     & +0.017 \upred     & +0.028 \upred     & +0.036 \upred     & +0.056 \upred     \\
NeuroLM  & +0.044 \upred & -0.012 \downgreen & -0.020 \downgreen & +0.019 \upred     & +0.001 \upred     & +0.037 \upred     \\
EEGPT    & +0.148 \upred & +0.159 \upred     & +0.122 \upred     & -0.003 \downgreen & -0.002 \downgreen & -0.005 \downgreen \\
\bottomrule
\end{tabular}

\vspace{0.65em}

\begin{tabular}{l ccc ccc}
\toprule
\multirow{2.5}{*}{\textbf{Model}} & \multicolumn{3}{c}{\textbf{BCI IV 2b}} & \multicolumn{3}{c}{\textbf{SEED-IV}} \\
\cmidrule(lr){2-4} \cmidrule(lr){5-7}
& $\Delta$BA & $\Delta$F1 & $\Delta$Kappa & $\Delta$BA & $\Delta$F1 & $\Delta$Kappa \\[0.5ex]
\midrule
NeuroGPT & +0.042 \upred & +0.044 \upred & +0.084 \upred & +0.042 \upred & +0.069 \upred     & +0.061 \upred \\
LaBraM   & +0.022 \upred & +0.017 \upred & +0.044 \upred & +0.013 \upred & -0.037 \downgreen & +0.014 \upred     \\
NeuroLM  & +0.086 \upred & +0.071 \upred & +0.172 \upred & +0.066 \upred & +0.167 \upred     & +0.083 \upred     \\
EEGPT    & +0.077 \upred & +0.046 \upred & +0.154 \upred & +0.128 \upred & +0.178 \upred     & +0.167 \upred     \\
\bottomrule
\end{tabular}
\end{table}

\begin{observationbox}{Observation 5 on No-GPT Ablation. } The contribution of the pretrained linguistic model in current FMs depends on both model architecture and domain shift. For classification-via-generation FMs (e.g., NeuroLM), removing the pretrained linguistic component causes prediction collapse. For direct classification FMs (e.g., NeuroGPT), the pretrained GPT component offers marginal impact on in-domain tasks, while its out-of-domain benefits remain task-dependent, yielding clear gains on BCI IV 2b but only marginal improvements on SEED-IV.     

\end{observationbox}

\textbf{No-GPT Ablation.} As shown in Table~\ref{Tab_GPT_Ablation}, the classification-via-generation FM (NeuroLM), which formulates prediction as conditional token generation (e.g., outputting ‘Yes/No’ for binary classification), suffers from prediction collapse (outputting empty strings) when the pretrained linguistic component is removed. This highlights the necessity of a pretrained linguistic backbone in this type of FM because linguistic knowledge guides the generation of valid textual class instructions. For the direct classification FM (NeuroGPT), which directly maps features to class probabilities, the out-of-domain benefits of the pretrained GPT are task-dependent, yielding gains of over 0.025 across all metrics on BCI IV 2b, but only marginal changes (up to 0.005) on SEED-IV. Meanwhile, it slightly degrades performance on in-domain tasks (e.g., TUEV and TUAB), with marginal decreases of 0.002–0.009 across metrics. Note that EEGPT and LaBraM do not include a pretrained GPT module in their architectures, so we do not conduct the No-GPT ablation for these models.  

\begin{table}[ht]
\centering
\scriptsize 
\setlength{\tabcolsep}{2.5pt} 
\renewcommand{\arraystretch}{1.15}
\caption{Performance gains from a pretrained linguistic component in FMs. $\Delta$ denotes the performance difference between FMs with a pretrained linguistic model and their ablated counterparts (Original $-$ Ablated), where positive values indicate gains from the linguistic component. The ablated NeuroLM fails to generate valid textual class instructions. Therefore, N/A* is used to report the performance difference. TUEV and TUAB are evaluated as in-domain tasks to maintain direct alignment with NeuroGPT's original TUH pretraining setup~\cite{cui2024neurogpt}.}
\label{Tab_GPT_Ablation}

\begin{tabular}{l ccc ccc}
\toprule
\multirow{2.5}{*}{\textbf{Model}} & \multicolumn{3}{c}{\textbf{TUEV}} & \multicolumn{3}{c}{\textbf{TUAB}} \\
\cmidrule(lr){2-4} \cmidrule(lr){5-7}
& $\Delta$BA & $\Delta$F1 & $\Delta$Kappa & $\Delta$BA & $\Delta$F1 & $\Delta$Kappa \\[0.5ex]
\midrule
NeuroGPT & -0.009 \downgreen & -0.009 \downgreen & -0.008 \downgreen & -0.002 \downgreen & -0.002 \downgreen & -0.004 \downgreen \\
NeuroLM  & N/A*              & N/A*              & N/A*              & N/A*              & N/A*              & N/A*              \\
\bottomrule
\end{tabular}

\vspace{0.65em}

\begin{tabular}{l ccc ccc}
\toprule
\multirow{2.5}{*}{\textbf{Model}} & \multicolumn{3}{c}{\textbf{BCI IV 2b}} & \multicolumn{3}{c}{\textbf{SEED-IV}} \\
\cmidrule(lr){2-4} \cmidrule(lr){5-7}
& $\Delta$BA & $\Delta$F1 & $\Delta$Kappa & $\Delta$BA & $\Delta$F1 & $\Delta$Kappa \\[0.5ex]
\midrule
NeuroGPT & +0.027 \upred & +0.029 \upred & +0.054 \upred & +0.004 \upred & \phantom{+}0.000 & +0.005 \upred \\
NeuroLM  & N/A*          & N/A*          & N/A*          & N/A*          & N/A*             & N/A*          \\
\bottomrule
\end{tabular}
\end{table}

\subsection{Neurophysiological Probing (NPP) Results} 
\label{Exp_NPPR}

\begin{table}[ht]
\centering
\scriptsize 
\setlength{\tabcolsep}{2.5pt} 
\renewcommand{\arraystretch}{1.15}
\caption{Performance changes induced by phase-randomized surrogate signals across all four benchmarks. $\Delta$ denotes the performance difference (Phase-randomized $-$ Original), where negative values indicate performance degradation.}
\label{Tab_NNP_PR}

\begin{tabular}{l ccc ccc}
\toprule
\multirow{2.5}{*}{\textbf{Model}} & \multicolumn{3}{c}{\textbf{TUEV}} & \multicolumn{3}{c}{\textbf{TUAB}} \\
\cmidrule(lr){2-4} \cmidrule(lr){5-7}
& $\Delta$BA & $\Delta$F1 & $\Delta$Kappa & $\Delta$BA & $\Delta$F1 & $\Delta$Kappa \\[0.5ex]
\midrule
NeuroGPT & -0.160 \downgreen & -0.163 \downgreen & -0.300 \downgreen & -0.064 \downgreen & -0.065 \downgreen & -0.128 \downgreen \\
LaBraM   & -0.117 \downgreen & -0.078 \downgreen & -0.112 \downgreen & -0.023 \downgreen & -0.035 \downgreen & -0.040 \downgreen \\
NeuroLM  & -0.116 \downgreen & -0.103 \downgreen & -0.110 \downgreen & -0.028 \downgreen & -0.051 \downgreen & -0.050 \downgreen \\
EEGPT    & -0.128 \downgreen & -0.112 \downgreen & -0.122 \downgreen & -0.058 \downgreen & -0.060 \downgreen & -0.112 \downgreen \\
\bottomrule
\end{tabular}

\vspace{0.65em}

\begin{tabular}{l ccc ccc}
\toprule
\multirow{2.5}{*}{\textbf{Model}} & \multicolumn{3}{c}{\textbf{BCI IV 2b}} & \multicolumn{3}{c}{\textbf{SEED-IV}} \\
\cmidrule(lr){2-4} \cmidrule(lr){5-7}
& $\Delta$BA & $\Delta$F1 & $\Delta$Kappa & $\Delta$BA & $\Delta$F1 & $\Delta$Kappa \\[0.5ex]
\midrule
NeuroGPT & -0.146 \downgreen & -0.174 \downgreen & -0.290 \downgreen & -0.075 \downgreen & -0.102 \downgreen & -0.099 \downgreen \\
LaBraM   & -0.036 \downgreen & -0.044 \downgreen & -0.072 \downgreen & -0.010 \downgreen & -0.006 \downgreen & -0.013 \downgreen \\
NeuroLM  & -0.004 \downgreen & -0.011 \downgreen & -0.010 \downgreen & -0.013 \downgreen & -0.018 \downgreen & -0.016 \downgreen \\
EEGPT    & -0.032 \downgreen & -0.018 \downgreen & -0.063 \downgreen & -0.086 \downgreen & -0.110 \downgreen & -0.112 \downgreen \\
\bottomrule
\end{tabular}
\end{table}

To verify EEG-FMs' neurophysiological feature dependencies, we utilize our NPP framework, which quantifies their reliance on specific physiological features through a comparative analysis of original EEG signals and systematically generated surrogate signals (i.e., phase-randomized, spatially perturbed, and spectrally ablated data).

\begin{observationbox}{Observation 6 on phase-randomization analysis.} Phase-randomized surrogate signals consistently reduce all metrics across all evaluated FMs and datasets. Continuous-representation models appear to rely more heavily on phase information for EEG decoding.

\end{observationbox}

The observed performance degradation across all models and datasets (see Table~\ref{Tab_NNP_PR}) demonstrates that these FMs utilize temporal phase-locked structures for predictions, though the magnitude of reliance varies significantly across model architectures and datasets. Our results show that continuous-representation models (NeuroGPT and EEGPT) exhibit greater performance degradation than discrete-representation models (LaBraM and NeuroLM) for most cases. This highlights that continuous-representation models are more dependent on temporal phase-locked structures when making predictions.

\begin{figure}[t]
\centering
\includegraphics[height=16cm ,width=20cm,angle=0,scale=0.4]{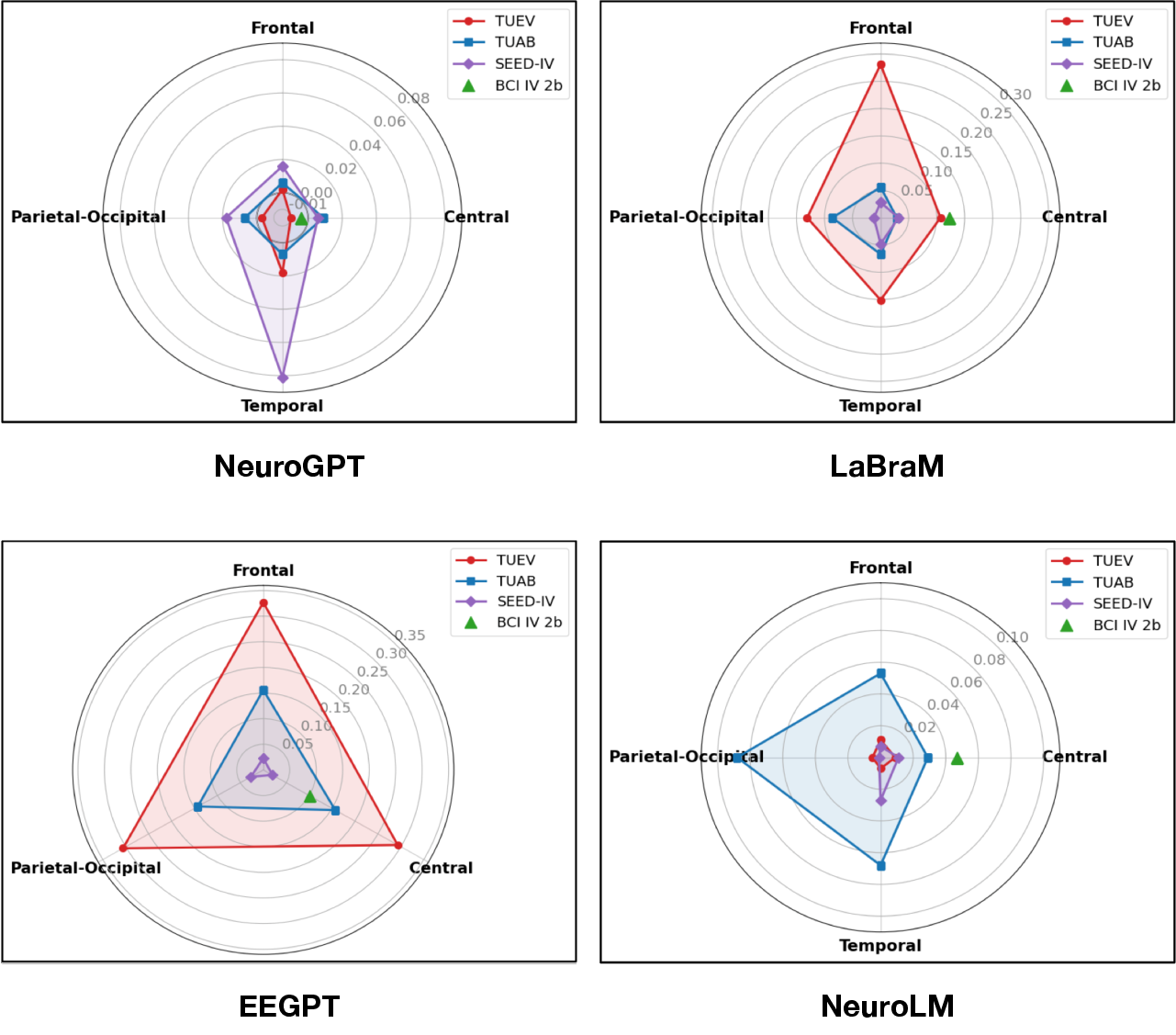}
\caption{Radar plots for spatial perturbation analysis on FMs. Radar plots show the contribution of specific brain regions to the decision-making of FMs by measuring performance variation under signal-aware Gaussian noise injection. Note that EEGPT excludes temporal regions due to pretraining channel misalignment on TUAB, TUEV and SEED-IV. BCI IV 2b is restricted to the central region due to its sparse 3-channel montage.}
\label{fig_SP_Results}
\end{figure}

\begin{observationbox}{Observation 7 on spatial perturbation analysis.} FMs correctly focus on task-relevant areas, though they show varying levels of sensitivity and regional weighting.

\end{observationbox}

As shown in Fig.~\ref{fig_SP_Results}, FMs exhibit task-congruent regional focus, aligning with neuroscientific findings. For example, injecting Gaussian noise ($\lambda = 0.5$, scaled by the batch standard deviation) into the EEG channels of the frontal and temporal lobes causes a substantial performance drop for clinical event classification (TUEV), indicating FMs primarily focus on these regions for the task. This is highly consistent with clinical literature identifying the frontal and temporal lobes as critical in epilepsy~\cite{vismer2015piriform,noebels2024jaspers}. Similarly, the experiments on SEED-IV show that perturbing the temporal lobe severely degrades most models. This is consistent with literature showing the critical role of lateral temporal channels in emotion decoding on SEED-IV~\cite{zheng2015investigating,zheng2019identifying}. Furthermore, we observe varying degrees of spatial reliance across models. Some FMs, such as EEGPT, show high absolute sensitivity values (up to 0.3) on TUEV compared to NeuroLM ($<$0.02), suggesting they depend more heavily on localized spatial features for predictions. Controlled experiments comparing targeted ROI perturbations against random channel corruption further distinguish neurophysiological sensitivity from generic distribution shift (see Section~\ref{app:npp_control} and Table~\ref{tab:spatial_control}).

\begin{figure}[t]
\centering
\includegraphics[height=16cm ,width=20cm,angle=0,scale=0.4]{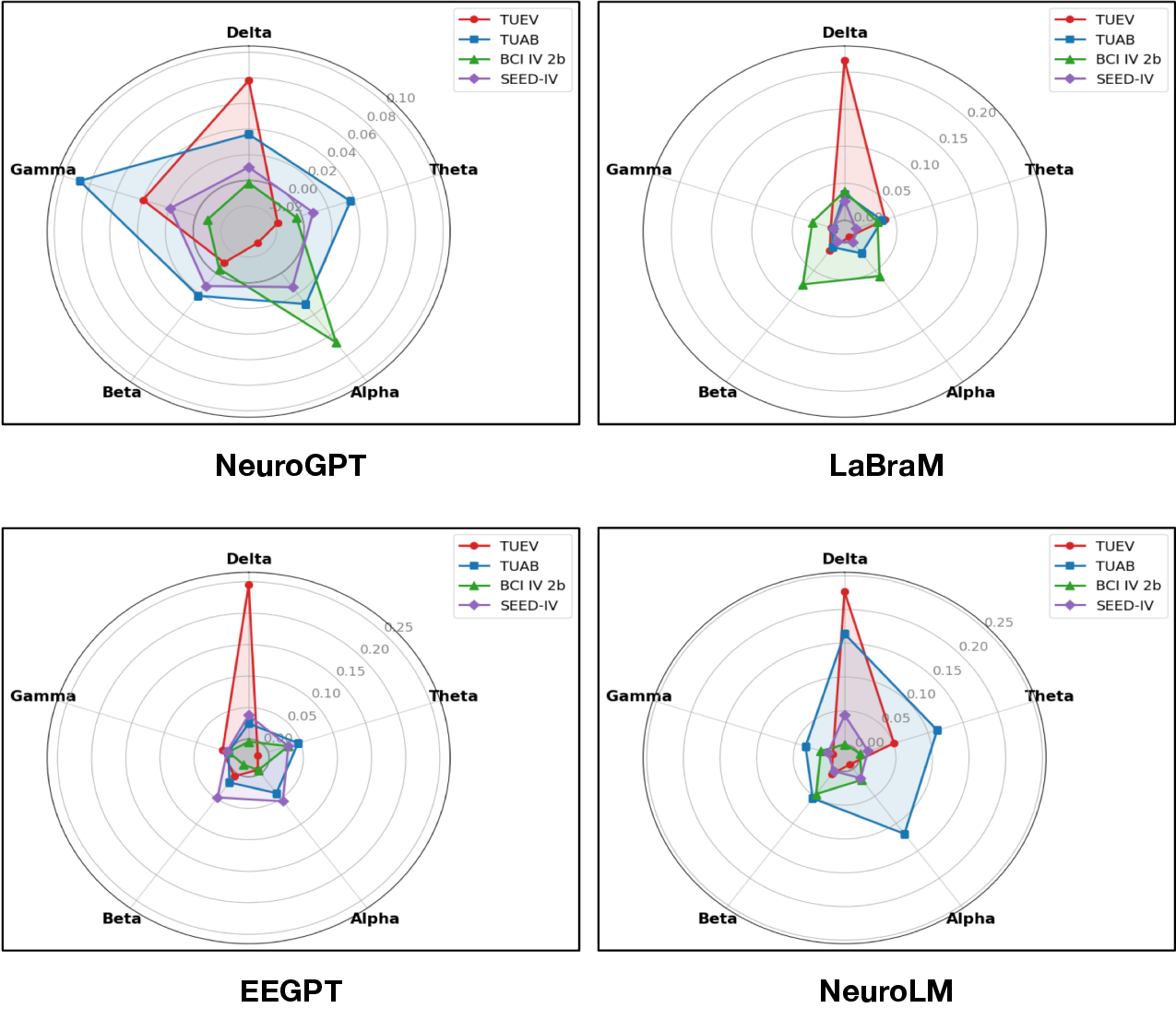}
\caption{Radar plots for spectral ablation analysis on FMs. Radar plots show the contribution of specific frequency bands to the decision-making of FMs by measuring performance variation under the removal of the specific frequency band. }
\label{fig_SA_results}
\end{figure}

\begin{observationbox}{Observation 8 on spectral ablation analysis.} FMs exhibit task-specific spectral sensitivity aligning with established neuroscientific findings. 
\end{observationbox}

Our results show that all evaluated FMs rely heavily on the Delta frequency band for clinical event classification (TUEV), consistent with recent clinical literature identifying slow-wave activity as a critical biomarker in epilepsy~\cite{pant2026exploring}. Similarly, several FMs generally focus on Alpha and Beta rhythms, aligning with well-documented event-related desynchronization (ERD)~\cite{pfurtscheller1994erd} associated with motor imagery. For FMs (e.g., EEGPT), they show high absolute sensitivity values ($>$ 0.2) on TUEV compared to NeuroGPT ($<$ 0.1), indicating that they depend more heavily on task-specific frequency bands for predictions. We observe similar task-specific frequency reliance in affective decoding on SEED-IV. Models using discrete tokenization (LaBraM and NeuroLM) exhibit high sensitivity to Delta-band perturbations, whereas continuous-representation models show broader spectral sensitivity. EEGPT is most sensitive to the Beta and Alpha bands, while NeuroGPT shows peak sensitivity in the Gamma band. This spectral pattern is consistent with prior affective EEG studies showing that Beta and Gamma oscillations are strongly associated with emotion recognition~\cite{zheng2015investigating,zheng2019identifying}. The stronger sensitivity of EEGPT and NeuroGPT to these higher-frequency bands also coincides with their relatively better downstream decoding performance. A systematic equal-width ablation sweep further indicates that degradation is band-specific rather than driven by generic filtering artifacts (see Section~\ref{app:npp_control} and Table~\ref{tab:spectral_control}).

\subsection{Assessing Neurophysiological Sensitivity against Distribution Shift}
\label{app:npp_control}

A potential concern regarding our neurophysiological probing (NPP) is whether observed performance degradation reflects sensitivity to genuine task-relevant neural patterns or simply arises from generic distribution shift. For example, frequency-domain perturbations can introduce spectral artifacts that may confound the interpretation of band-specific performance changes~\cite{wang2026fouriermask}. To further assess these effects, we conducted controlled experiments on TUEV evaluating ROI perturbation and equal-width spectral ablation  against matched controls across two representative EEG-FMs (NeuroGPT and LaBraM) and two supervised baselines (MSCFormer and TSSEFFNet).

\paragraph{Targeted vs. Random Spatial Perturbation}
To examine spatial sensitivity, we compared targeted ROI perturbations (temporal channels for NeuroGPT and MSCFormer; frontal channels for LaBraM and TS-SEFFNet) with channel-matched random perturbations under two noise levels ($\mathrm{NL} \in \{0.5, 2.0\}$). As shown in Table~\ref{tab:spatial_control}, under mild noise ($\mathrm{NL}=0.5$), targeted ROI perturbation consistently induces a statistically significant drop relative to the unperturbed baseline and produces larger performance degradation than perturbing an equal number of randomly selected channels. Under severe corruption ($\mathrm{NL}=2.0$), performance degradation becomes more comparable between targeted and random perturbations, consistent with the broader impact of substantial signal corruption. The stronger sensitivity to mild targeted noise provides evidence consistent with models relying preferentially on anatomically relevant spatial regions for clinical event classification rather than being uniformly sensitive across all channels.

\paragraph{Equal-Width Spectral Ablation Sweep}
We performed a systematic 4~Hz band-stop ablation sweep on TUEV from 0.1 to 72~Hz (18 bins: 0.1--4~Hz, 4--8~Hz, \ldots, 68--72~Hz) to determine whether performance degradation is specific to particular frequency bands rather than an artifact of spectral filtering. Because DC offset and slow drift below 0.1~Hz were removed during preprocessing (0.1--75~Hz bandpass), the significant degradation ($p < 0.05$, paired two-tailed $t$-test) concentrated in the 0.1--4~Hz Delta band (Table~\ref{tab:spectral_control}) reflects genuine reliance on low-frequency neural features, consistent with the expected prominence of slow-wave activity in epileptic TUEV recordings. In contrast, higher-frequency bins, including the 32--36~Hz control band, generally produce small and non-significant performance changes. These results suggest that the observed performance degradation is driven primarily by the removal of low-frequency EEG activity rather than by generic distortion introduced by the filtering procedure.

\begin{table}[ht]
\centering
\scriptsize
\setlength{\tabcolsep}{4pt}
\renewcommand{\arraystretch}{1.12}
\caption{Targeted vs. random spatial perturbation on TUEV (5 random seeds). Values denote percentage-point changes in Balanced Accuracy ($\Delta\mathrm{BA} = (\mathrm{BA}_{\mathrm{perturbed}} - \mathrm{BA}_{\mathrm{original}}) \times 100$, in percentage points) relative to the unperturbed baseline. ROI denotes temporal channels for NeuroGPT and MSCFormer, and frontal channels for LaBraM and TS-SEFFNet. Asterisks ($*$) denote statistical significance ($p < 0.05$, paired two-tailed $t$-test against unperturbed baseline).}
\label{tab:spatial_control}
\begin{tabular}{lcccc}
\toprule
\textbf{Configuration} & \textbf{NeuroGPT} & \textbf{LaBraM} & \textbf{MSCFormer} & \textbf{TS-SEFFNet} \\
\midrule
ROI ($\mathrm{NL} = 0.5$)     & $-1.77^*$          & $-27.54^*$ & $-3.55^*$          & $-5.01^*$          \\
Random ($\mathrm{NL} = 0.5$)  & $-0.66\phantom{*}$ & $-9.57^*$  & $-1.26\phantom{*}$ & $-0.95\phantom{*}$ \\
\midrule
ROI ($\mathrm{NL} = 2.0$)     & $-19.30^*$         & $-25.33^*$ & $-9.70^*$          & $-20.98^*$         \\
Random ($\mathrm{NL} = 2.0$)  & $-18.69^*$         & $-22.71^*$ & $-6.31^*$          & $-14.18^*$         \\
\bottomrule
\end{tabular}
\end{table}

\begin{table}[ht]
\centering
\scriptsize
\setlength{\tabcolsep}{4pt}
\renewcommand{\arraystretch}{1.12}
\caption{Equal-width frequency ablation sweep on TUEV (5 random seeds). Continuous 4\,Hz band ablation (0.1--72\,Hz, 18 bins) evaluating percentage shifts in Balanced Accuracy ($\Delta\mathrm{BA} = (\mathrm{BA}_{\mathrm{ablated}} - \mathrm{BA}_{\mathrm{original}}) \times 100$, in percentage points). Asterisks ($*$) denote statistical significance ($p < 0.05$, paired two-tailed $t$-test). Unlisted bins exhibit only marginal, non-significant changes ($-1\%$ to $+1\%$).}
\label{tab:spectral_control}
\begin{tabular}{lcccc}
\toprule
\textbf{Frequency Band} & \textbf{NeuroGPT} & \textbf{LaBraM} & \textbf{MSCFormer} & \textbf{TS-SEFFNet} \\
\midrule
0.1--4\,Hz (Delta)     & $-8.81^*$           & $-13.67^*$ & $-24.87^*$ & $-16.86^*$ \\
4--8\,Hz (Theta)     & $+1.05\phantom{*}$  & $-2.59^*$  & ---        & $-5.12^*$  \\
8--12\,Hz (Alpha)    & $+2.93^*$           & ---        & ---        & ---        \\
32--36\,Hz (Control) & $-1.15^*$           & ---        & ---        & ---        \\
\bottomrule
\end{tabular}
\end{table}

\subsection{Rationale for Asymmetric Optimization Strategies}
\label{app:fm_AOS}

We illustrate why we adopt different optimization strategies to benchmark supervised baseline models and EEG foundation models here. For supervised models, they are generally trained from scratch, so their performance is usually sensitive to hyperparameter selection (see Fig~\ref{Fig_ASHA_Effectiveness}). Thus, careful hyperparameter optimization is necessary when adopting them to different datasets. To this end, we use an ASHA-driven hyperparameter optimization strategy for supervised models, leveraging ASHA's early-stopping mechanism to efficiently identify best-performing architectural and optimization hyperparameters from a large number of candidate configurations within a reasonable search time. For EEG foundation models, whose pretrained architectures are fixed, we perform grid search over their limited set of optimization hyperparameters. This avoids the potential risk of prematurely eliminating configurations through early stopping, ensures complete coverage of their relatively small search space, and circumvents the prohibitive computational overhead of running multiple EEG-FM trials concurrently with ASHA-driven hyperparameter optimization.

Note that optimizing the pretraining stage could yield improved downstream performance for EEG-FMs. However, performing such optimization for each model and dataset would require substantial additional pretraining (e.g., the pre-training dataset for NeuroGPT had $\sim$5656 hours of data and $\sim$25000 hours for NeuroLM), while improvements in pretraining loss cannot ensure downstream performance must be improved due to potential misalignment between the pretraining objective and downstream tasks. Therefore, we do not optimize the pretraining stage in this study. Instead, we directly use the pretrained models released by their original authors.

\section{Conclusion}
In this paper, we introduced EEG-FM-Audit, a comprehensive evaluation and analysis pipeline for EEG-FMs that ensures transparent and fair benchmarking, effective assessment of learning paradigms, and rigorous verification of models' reliance on neurophysiological features. By evaluating four tuned leading FMs against tuned supervised baselines, we found that FMs do not yet hold a definitive advantage in EEG classification, particularly on small downstream datasets, and that the effectiveness of their learning paradigms depends strongly on dataset scale and model architecture. Importantly, our analysis reveals that current FMs capture EEG properties that are consistent with established neuroscientific findings. Through the EEG-FM-Audit pipeline, we provide the community with a rigorous evaluative framework to directly inform the development of more robust and physiologically consistent EEG foundation models.

Despite these contributions, this study has several limitations that motivate key directions for future work. First, while our experiments cover four representative datasets and nine architectures, broader evaluation across additional clinical paradigms, model families, and tuning strategies is needed to further assess generalizability. Second, although we replace the pretrained linguistic backbone in certain EEG-FMs with three randomly initialized self-attention layers to isolate its impact on EEG decoding, alternative pretrained non-linguistic backbones (e.g., vision or audio Transformers) warrant further investigation. Third, the benchmarking process introduces substantial computational overhead from repeated hyperparameter optimization and neurophysiological probing; future work will explore more computationally efficient auditing protocols.



\bibliographystyle{IEEEtran}
\bibliography{references}

\vfill

\end{document}